\documentclass{article}

\usepackage[preprint]{neurips_2024}

\usepackage[utf8]{inputenc} 
\usepackage[T1]{fontenc}    
\usepackage{hyperref}       
\usepackage{url}            
\usepackage{booktabs}       
\usepackage{amsfonts}       
\usepackage{nicefrac}       
\usepackage{microtype}      
\usepackage[dvipsnames, svgnames, x11names]{xcolor}
\definecolor{aliceblue}{rgb}{0.94, 0.97, 1.0}
\definecolor{deeppink}{RGB}{255,20,147}
\definecolor{mygray}{gray}{0.9}

\usepackage{graphicx}
\usepackage{wrapfig}
\usepackage{subfigure}
\usepackage{multirow}
\usepackage{amsmath}
\usepackage{mathtools}
\usepackage{mathrsfs}
\usepackage{amsthm}
\usepackage{verbatim}

\usepackage{amsmath,amsfonts,bm}

\def\eqref#1{equation~\ref{#1}}

\def\1{\bm{1}}

\def\va{{\bm{a}}}

\def\vy{{\bm{y}}}

\def\mI{{\bm{I}}}

\def\mK{{\bm{K}}}

\def\mQ{{\bm{Q}}}

\def\mS{{\bm{S}}}

\def\mV{{\bm{V}}}
\def\mW{{\bm{W}}}
\def\mX{{\bm{X}}}
\def\mY{{\bm{Y}}}

\DeclareMathAlphabet{\mathsfit}{\encodingdefault}{\sfdefault}{m}{sl}
\SetMathAlphabet{\mathsfit}{bold}{\encodingdefault}{\sfdefault}{bx}{n}

\newcommand{\R}{\mathbb{R}}

\title{Prompt-Aware Adapter: Towards Learning Adaptive Visual Tokens for Multimodal Large Language Models}

\author{%
  Yue~Zhang~~~~~~~~~~~~~~~~~~~~~~Hehe~Fan~~~~~~~~~~~~~~~~~~~~~~Yi~Yang \\
  School of Computer Science and Technology\\ 
  Zhejiang University\\
}

\begin{document}

\maketitle

\begin{abstract}
To bridge the gap between vision and language modalities, Multimodal Large Language Models (MLLMs) usually learn an adapter that converts visual inputs to understandable tokens for Large Language Models (LLMs). 
However, most adapters generate consistent visual tokens, regardless of the specific objects of interest mentioned in the prompt. 
Since these adapters distribute equal attention to every detail in the image and focus on the entire scene, they may increase the cognitive load for LLMs, particularly when processing complex scenes. 
To alleviate this problem, we propose prompt-aware adapters. 
These adapters are designed with the capability to dynamically embed visual inputs based on the specific focus of the prompt.
Specifically, prompt-aware adapters utilize both global and local textual features to capture the most relevant visual clues from the prompt at both coarse and fine granularity levels.
This approach significantly enhances the ability of LLMs to understand and interpret visual content. 
Experiments on various visual question answering tasks, such as counting and position reasoning, demonstrate the effectiveness of prompt-aware adapters. 
\end{abstract}

\section{Introduction}\label{sec:intro}
Recent advances in Large Language Models (LLMs)~\cite{brown2020language, chung2022scaling, openai2022chatgpt, touvron2023llama, chiang2023vicuna} have significantly enhanced their performance across various natural language processing tasks. 
These models are able to perform language comprehension and logical reasoning, enabling them to handle complex linguistic functions such as summarizing texts, answering questions, processing dialogues, and composing new essays or articles. 
However, LLMs are inherently limited by their inability to process visual information. 
This has led to the development of Multimodal Large Language Models (MLLMs)~\cite{driess2023palm, yang2023mm, li2023videochat, doveh2024dense, peng2023kosmos}, which expand upon LLMs by incorporating visual processing abilities. 
MLLMs offer a more holistic understanding and interaction by synthesizing both textual and visual data, thereby broadening their utility in diverse real-world applications.


\begin{figure}
    \centering
    \includegraphics[width=1.0\textwidth]{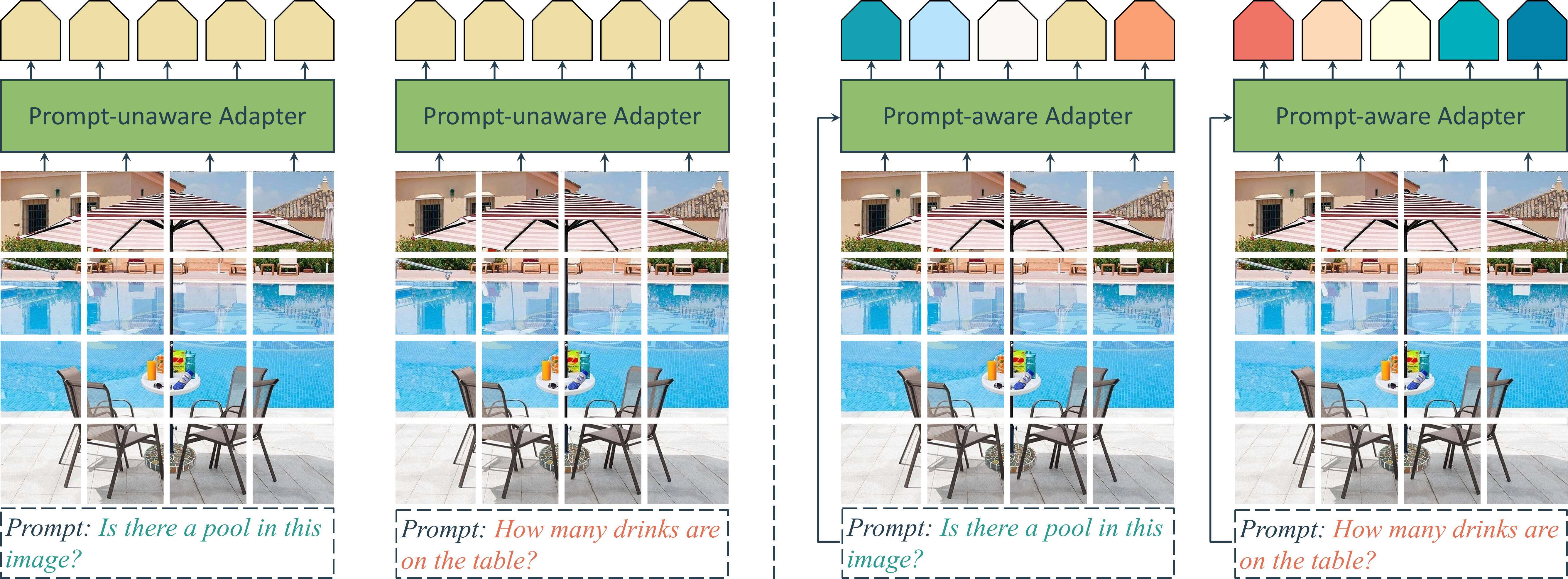}
    \caption{Illustration comparing prompt-unaware and prompt-aware adapters. \textbf{Left:} Prompt-unaware adapter treats visual patches as a kind of words and directly converts these patches into ``readable'' tokens for LLMs, without considering the specific objects of interest. In this case, whether the question involves ``pool'' or ``drinks,'' they consistently generate the same tokens and allocate equal attention to every detail in the scene, which may increase the cognitive load for LLMs.
    \textbf{Right:} Prompt-aware adapter leverages prompts to collect the most relevant visual clues and generate adaptive tokes, thus enhancing the ability of LLMs to understand and interpret visual content. 
    }
    \label{fig:intro}
\end{figure}

To equip LLMs with visual perception ability, MLLMs typically employ a trainable adapter that connects a frozen visual encoder and a frozen LLM.
Adapters play a crucial role in bridging the gap between vision and language, enabling profound visual understanding while leveraging the powerful reasoning capabilities of LLMs. 
However, most existing adapters~\cite{liu2023visual,chen2023minigpt,chen2023shikra,zeng2023matters, su2023pandagpt, peng2023kosmos, zhou2023infmllm,cha2023honeybee} approach visual patches as if they were words, and directly translate these patches into tokens comprehensible to LLMs (e.g., through linear projection), regardless of the specific objects of interest in the prompt. 
These prompt-unaware adapters may enable LLMs to correctly analyze  simple images but could struggle to understand complex scenes. 
As shown in Fig.~\ref{fig:intro}, no matter the query is ``is there a pool in this image'' or ``how many drinks are on the table'', these adapters consistently convert the image into the same tokens. 
Consequently, the subsequent LLMs have to independently parse the scene to deduce the spatial context and
shift attention to  ``pool'' or ``drinks'' entirely on their own.

A few studies have explored using prompts to guide the behavior of adapters. 
For instance, VisionLLM~\cite{wang2023visionllm} and Flamingo~\cite{alayrac2022flamingo} employs the cross-attention mechanism to learn adaptive visual tokens, where prompt words serve as the query and image patches are treated as the key and value. 
InstructBLIP~\cite{dai2024instructblip} first injects the prompt information into learned queries~\cite{li2023blip} (via self-attention) and then uses cross-attention to gather visual clues. 
These cross-attention-based adapters encounter two main challenges. 
First, they search for visual clues at the word level, thereby neglecting global information for capturing an overview of the prompt-related region. 
Second, the use of the softmax function in cross-attention normalizes attention distribution from word to patch such that the total attention allocated to each word equals 1. 
This implies that every word in the prompt, including function words like ``a'', ``the'' and ``is'', is forced to correspond to a specific region in the image.
Because these irrelevant words may correspond to different regions each time, this unrealistic assumption may lead adapters to produce unstable visual tokens, inevitably causing significant confusion for LLMs. 

In this paper, we produce a prompt-aware adapter to adaptively embed visual inputs based on the prompt's global and local representations. 
Our adapter comprises two key components: prompt-aware global attention and prompt-aware local attention. 
The global attention is designed to capture coarse-grained, prompt-related visual perceptions, while the local attention focuses on refining responses to specific, fine-grained regions of interest. 
This dual approach allows adapters to effectively uncover visual contexts and shift attention to relevant areas as needed.
We conducted extensive experiments on the COCO-QA~\cite{ren2015exploring} and MME~\cite{fu2023mme} datasets. 
Compared to the prompt-unaware baseline, our method shows significant improvements on COCO-QA, enhancing object classification, counting, color recognition, and positional reasoning by 7.71\%, 18.42\%, 12.84\%, and 9.51\%, respectively. On the MME dataset, it boosts the total scores for perception tasks and cognition tasks by 59.43\% and 46.91\%, respectively. 
Our method demonstrates superior performance in handling complex scenarios and parsing intricate problems, effectively sensing prompts and capturing informative details necessary for answering questions. 
The contributions are as follows:
\vspace{-0.5em}
\begin{itemize}
    \item 
    Among the early efforts, we conduct a comprehensive study on the impact of prompts on adapters. 
    Our research reveals that prompt-independent adapters may be not sufficient to capture the most informative visual clues for visual understanding. 
    Furthermore,  existing prompt-relevant adapters still suffer from issues such as failing to focus on crucial words in prompts or adequately aligning visual and semantic elements. 
    \item 
    We propose prompt-aware adapters that dynamically embed visual inputs based on prompts at both coarse and fine granularities. 
    Our method provides effective and convenient attention mechanisms for existing adapter-empowered MLLMs.
    \item 
    Experiments on complex scene understanding applications demonstrate that the proposed method effectively improves the visual perception and reasoning capabilities of MLLMs. 
\end{itemize}

\section{Related Work}\label{sec:rw}
\textbf{Multimodal Large Language Models and Adapters.} 
The integration of the perceptual abilities of vision models~\cite{kirillov2023segment, mokady2021clipcap, oquab2023dinov2, radford2021learning} with the reasoning capabilities of LLMs~\cite{brown2020language, ouyang2022training, chowdhery2022palm, thoppilan2022lamda, zhang2022opt, openai2022chatgpt} has given rise to Multimodal Large Language Models (MLLMs)~\cite{yang2024doraemongpt, hu2024minicpm, chen2023sharegpt4v, zheng2024dreamlip, li2024mini, li2024cir}. 
In MLLMs, visual signals are converted into tokens that LLMs can understand, typically using adapters. 
For example, MLLMs such as LLaVA~\cite{liu2023visual}, Shikra~\cite{chen2023shikra},  MiniGPT~\cite{chen2023minigpt}, Matters~\cite{zeng2023matters}, PandaGPT~\cite{su2023pandagpt}, Kosmos~\cite{peng2023kosmos}, and InfMLLM~\cite{zhou2023infmllm} utilize linear projection adapters.  
QWEN-VL~\cite{bai2023qwen} and mPlUG-Owl~\cite{ye2023mplug} employ Perceiver-like architectures~\cite{jaegle2021perceiver} as adapters. 
Since the introduction of the Q-Former in BLIP-2~\cite{li2023blip}, many MLLMs~\cite{zhu2023minigpt, yu2023reformulating} have adopted it as an adapter for modal alignment. 
The visual descriptions generated by the aforementioned prompt-independent adapters typically encompass the entire image, without emphasizing the specific visual details relevant to the prompt. 
VisionLLM~\cite{wang2023visionllm}, Flamingo~\cite{alayrac2022flamingo}, and subsequent studies~\cite{gong2023multimodal, li2023otter} utilize cross-attention adapters to improve prompt-aware capabilities. 
InstructBLIP~\cite{dai2024instructblip} and Cheetor~\cite{li2023fine} incorporate prompt information into learned queries using self-attention and then employ cross-attention to collect visual clues.
These adapters often struggle with issues such as failing to focus on crucial words in prompts or adequately aligning visual and semantic elements. 
In this paper, we introduce a new type of adapter designed to align visual signals with textual semantics for MLLMs.

\textbf{Attention Mechanisms.} 
The attention mechanism~\cite{DBLP:journals/corr/BahdanauCB14} enables vision-language models to focus on relevant information within the input while minimizing focus on irrelevant details. Its effectiveness and interpretability have inspired numerous subsequent studies~\cite{lu2023multi, zheng2023less, yao2023attention}. For example, \cite{xu2015show} introduces both hard and soft attention mechanisms that automatically learn to describe the contents of images. 
Additionally, in the Visual Question Answering (VQA) task, it is crucial to focus on the relevant parts of both the image and the question. 
Consequently, Co-Attention~\cite{lu2016hierarchical} was developed to facilitate simultaneous attention learning across multiple inputs (i.e., image and text). 
Self-Attention~\cite{yang2016hierarchical} enables interactions between different positions within a single input sequence, thereby capturing content that is relatively more important. Drawing from sources such as \cite{yang2016hierarchical, lin2017structured, rocktaschel2017frustratingly}, the Transformer~\cite{vaswani2017attention}, recognized as a foundational component of large models, incorporates Scaled Dot-Product Attention and Multi-Head Attention. Cross-Attention~\cite{wei2020multi} has been developed within the Transformer~\cite{vaswani2017attention} architecture to facilitate interaction between two different sequences. To enhance the interaction between images and text, we propose  global and local attention mechanisms that automatically focus on the visual content of interest mentioned in the text.

\section{Proposed Method}\label{sec:mothod}
In this section, We first provide a brief overview of existing adapters for the visual perception of Multimodal Large Language Models (MLLMs). 
Then, we describe how the proposed prompt-aware adapter enables MLLMs to effectively uncover visual context and adaptively shift attention for enhancing visual reasoning.

\subsection{Preliminary: MLLM and Adapter}

MLLM is an advanced type of artificial intelligence model that processes and understands information from multiple types of data, such as text, images, and sometimes audio or video. 
This allows the model to perform tasks that involve more than one mode of communication, providing a richer understanding of the content than the text-only LLM.
MLLMs integrate capabilities from different domains. 
For example, they can extract meaning from both the text in an image caption and the visual content of the image itself. 
This is particularly useful in applications such as image captioning, 
Visual Question Answering (VQA) and video understanding. 

In MLLMs, a specialized module called adapter is commonly employed to enhance the model's capability to process and integrate information from different modalities. 
Adapters are particularly valuable because they allow pre-trained models to be adapted to new tasks or modalities without the need for extensive retraining of the entire model. 
Formally, for a given visual input, a frozen visual encoder (such as ViT and Q-Former) is typically employed to extract the visual features $\mX \in \R^{N \times C}$, where $N$ represents the number of patches and $C$ denotes the number of feature channels. 
Because these visual features are not directly comprehensible to the subsequent LLM, the adapter is trained to translate them into tokens that the LLM can understand as follows,
\begin{equation}
    \mX' = \mathrm{Adapter}(\mX;\theta),
\end{equation}
where $\theta$ is the learnable parameters of the adapter, $\mX' \in \R^{N' \times C'}$ and $N'$ represents the number of converted tokens and $C'$ denotes the dimension of the LLM's input.
In most MLLMs, the adapter is implemented as a linear projection layer that projects visual features into the textual feature space~\cite{liu2023visual, chen2023minigpt, zeng2023matters, su2023pandagpt, chen2023shikra}. 
Additionally, the adapter in Honeybee~\cite{cha2023honeybee} first applies a convolution to convert visual features and then downsamples them, resulting in $N' < N$. 
However, no matter linear projection or convolution, these adapters  distribute equal attention to every detail in the visual input and focus on the entire scene, regardless of the specific objects of 
interest mentioned in the prompt, converting visual input to invariant tokens. 
In this case, these prompt-unaware adapters may struggle to effectively convert complex scenes for its subsequent LLM.

\begin{figure}
    \centering
    \includegraphics[width=1.0\textwidth]{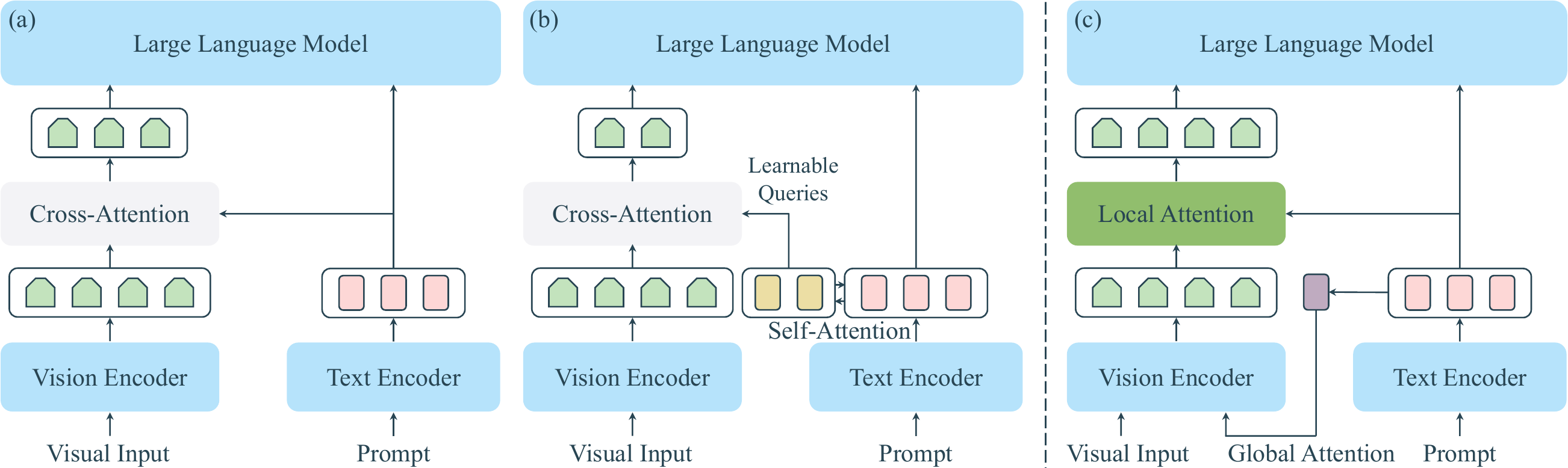}
    \caption{Illustration comparing the cross-attention (left) and proposed (right) adapters. 
    (a) Methods like VisionLLM~\cite{wang2023visionllm} and Flamingo~\cite{alayrac2022flamingo} utilize text features as queries and visual features as keys and values in cross-attention. 
    It assumes that each word in the prompt corresponds to specific regions. 
    The number of converted visual tokens is equal to that of text features. 
    (b) InstructBLIP~\cite{dai2024instructblip} first injects prompt information into learnable queries via self-attention, and then employs cross-attention. 
    It assumes that each query in the learnable queries corresponds to specific regions.   
    The number of converted visual tokens is equal to that of learnable queries. 
    (c) Our adapter comprises global and local attention components. Due to the new attention calculation mechanism used in local attention, the number of converted visual tokens remains unchanged. 
    }
    \label{fig:comp} 
\end{figure}

\subsection{Prompt-Aware Adapter}
To translate informative visual tokens for LLMs, we can use prompts to guide the behavior of adapters.
Suppose $\mY \in \R^{M \times D}$ is the prompt, where $M$ represents the number of words in the prompt and $D$ denotes the dimension of the word embedding. 
This mechanism can be formulated as follows, 
\begin{equation}
\label{eq:aware}
    \mX' = \mathrm{Adapter}(\mX, \mY; \theta). 
\end{equation}
There are many ways to implement Eq.~(\ref{eq:aware}).
As shown in Fig~\ref{fig:comp}, VisionLLM~\cite{wang2023visionllm} and Flamingo~\cite{alayrac2022flamingo} utilize text features as queries and visual features as keys and values in cross-attention processes to facilitate modal interaction. 
InstructBLIP~\cite{dai2024instructblip} first incorporates prompt information into learned queries through self-attention, and then employs cross-attention to collect visual clues. 
In summary, these methods are based on cross-attention, which can be formulated as follows,
\begin{equation}
\label{eq:cross}
    \mX' = \mathrm{softmax}(\frac{\mQ\cdot\mK^T}{\sqrt{E}})\cdot \mV,~~~~~~~\mQ=\mY\cdot\mW_q,~~\mK=\mX\cdot\mW_k,~~\mV=\mX\cdot\mW_v, 
\end{equation}
where $\mW_q \in \R^{D\times E}$, $\mW_k \in \R^{C\times E}$, $\mW_v \in \R^{C\times E}$ and $\cdot$ denotes matrix multiplication. 
In this paper, we use the lowercase $\mathrm{softmax}$ to indicate that the function is applied at each row. 
 This mechanism results in two outcomes:
\begin{itemize}
    \item  \textbf{Attention}. Because the use of the softmax function in cross-attention normalizes the attention distribution from each word to all patches, the total attention allocated to each word equals 1. Specifically, if $\va_{ij}$ represents the attention of the $i$-th word to the $j$-th patch, then $\sum_{j=1}^N \va_{ij}=1$. 
    \item \textbf{Output}. Because the prompt $\mY$ serves as the query in cross-attention, it leads to $\mX' \in \R^{M \times E}$. This indicates that the number of converted visual tokens is $M$ (the number of words in the prompt), rather than $N$ (the number of patches in the visual input). 
\end{itemize}
This suggests that every word in the prompt, including function words like ``a'', ``an'', ``the'', ``is'' and ``are'', is forced to correspond to a specific region in the image.
Given that these irrelevant words might correspond to different regions each time, this unrealistic assumption can cause adapters to generate unstable visual tokens, leading to significant confusion for LLMs.

\begin{figure}
    \centering
    \includegraphics[width=1.0\textwidth]{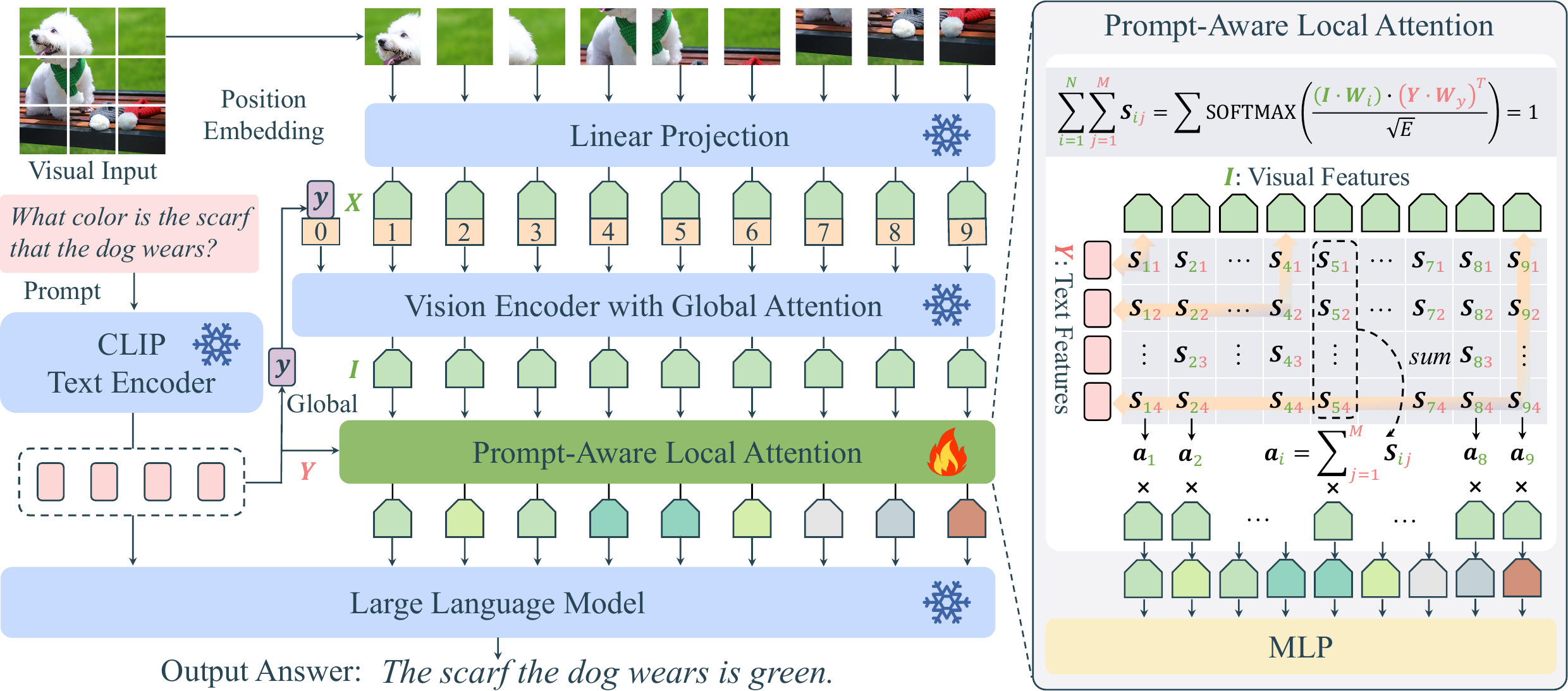}
    \caption{
Illustration of the proposed prompt-aware adapter. The adapter consists of a global attention component and a local attention component. The global attention, integrated into the visual encoder, is designed to capture coarse-grained, prompt-related visual perceptions. Meanwhile, the local attention focuses on refining responses to specific, fine-grained regions of interest.
    }
    \label{fig:arch} 
\end{figure}
\textbf{Global Attention.} To extract the most informative clues, we propose the prompt-aware adapter (shown in Fig.~\ref{fig:arch}). 
The adapter consists of a prompt-aware global attention component and a prompt-aware local attention component.  
The global attention component searches for visual clues at the scene level, aiming to capture an overview of the region related to the prompt. 
To this end, we first utilize CLIP's text encoder to extract the global feature of the prompt.  
Second, we employ a learned projection layer to map the global feature into the same space as the visual patch, i.e., $\vy \in \R^{1 \times C}$. 
Third, the global feature of the prompt is appended to the visual patches, resulting in $[\mX; y] \in \mathbb{R}^{(N+1) \times C}$. 
Fourth, self-attention is employed to allow the adapter to incorporate the global prompt information, which is as follows, 
\begin{equation}
    \mI = \mathrm{SelfAttention}([\vy; \mX]). 
\end{equation}
In this way, a coarse extraction of visual clues mentioned in the prompt is conducted. 
Finally, the global attention discards the prompt representation at the last position, yielding $N$ visual features, i.e., $\mI \in \R^{N\times C_i}$, where $C_i$ is the feature dimension. 

\textbf{Local Attention.}  Global attention reflects the overall correlation between the visual input and prompt. 
To capture local details, we design a local attention component. 
First, local attention calculates a similarity matrix $\mS \in \R^{N \times M}$ between text and visual features, as follows, 
\begin{equation}
\label{eq:local}
    \mS = \mathrm{SOFTMAX}\Big(\frac{\big(\mI\cdot\mW_i)\cdot(\mY\cdot\mW_y)^T}{\sqrt{E}}\Big)
\end{equation}
where $\mW_i \in \R^{C_i\times E}$ and $\mW_y \in \R^{D\times E}$. 
Here, we use the capital $\mathrm{SOFTMAX}$ to denote that the function is applied to the entire matrix. 
Compared to the $\mathrm{softmax}$ function used in Eq.~(\ref{eq:cross}), $\mathrm{SOFTMAX}$ suggests that each word may or may not correspond to a visual patch in the scene, and similarly, each patch may or may not correspond to a word in the prompt. 
This assumption is more flexible and realistic than that of cross-attention. 

Second, the sum of each row ($\mathbb{R}^{1\times M}$) in $\mS$ can be interpreted as the correlation between a specific patch and the entire prompt description.
Therefore, we calculate the weight of each visual patch by summing all attention focused on it, as follows, 
\begin{equation}
    \va_i = \sum_{j=1}^M \mS_{ij},~~~\mX' = \mathrm{MLP}(\va^T\cdot\mI), 
\end{equation}
where $\va \in \R^{1 \times N}$. 
Then, local attention is applied to visual features. This process produces prompt-aware visual tokens. Finally, a multilayer perceptron (MLP) is employed to transform them. 

Our method intrinsically differs from the cross-attention mechanism described in Eq.~(\ref{eq:cross}). 
In cross-attention, prompt features $\mY$ function as $\mQ$ (queries) and visual features $\mX$ as $\mK$ (keys) and $\mV$ (values). 
Here, each text token distributes its attention across visual patches such that the total sums to 1. 
This means that even function words (like conjunctions and prepositions) must have their similarity scores with all visual tokens collectively equal 1. 
This requirement is impractical, as it assumes that every text token is equally significant in directing attention to visual cues.
The proposed local attention mechanism ensures that $\sum_{i=1}^N\sum_{j=1}^M\mS_{i,j} = 1$. As a result, text tokens that are more relevant to the visual context (and have higher similarity scores) exert greater influence on visual encoding. This approach, akin to emphasizing certain words in a sentence when reading aloud, allows for the extraction of prompt-related visual features in a more detailed and nuanced manner. 
In summary, there are two differences between cross-attention and the proposed local attention.
\begin{itemize}
    \item \textbf{Attention.} The cross-attention implies that each word corresponds to a specific region, whereas local attention does not enforce such a correspondence. 
    \item \textbf{Output.} The cross-attention produces $M$ (the number of words in the prompt) visual tokens, whereas local attention generates $N$ (the number of patches in the visual input)  tokens. 
\end{itemize}

\section{Experiments}\label{sec:exper}
\subsection{Implementation Details}\label{sec:details}
\textbf{Network Details.}
In this paper, we utilize the open-sourced LLaMA2 (7B) model~\cite{touvron2023llama2} as our Large Language Model (LLM). We employ the text encoder from CLIP~\cite{radford2021learning} to ensure that the extracted text features closely align with the corresponding visual features in the embedding space. 
ViT-g/14 from EVA-CLIP~\cite{fang2023eva} is used as the vision encoder. 
The global text token is injected into the vision encoder to  conduct prompt-aware global attention. 
The use AdamW ~\cite{loshchilov2017decoupled} as the optimizer, with $\beta_1 = 0.9$, $\beta_2 = 0.999$, and a weight decay rate of $0.05$. 
The learning rate is linearly warmed up from $10^{-6}$ to $8 \times 10^{-5}$ over the first $1,000$ steps to accelerate model convergence. This is followed by a cosine decay of the learning rate to a minimum of $10^{-5}$. The training is set to a maximum of $50$ epochs, with $1,000$ iterations per epoch. 
The model is trained with a batch size of 4, across a period of 3 days on a single NVIDIA RTX A6000 GPU.  

\textbf{Training Details and Datasets.}
Following~\cite{touvron2023llama2,chen2023minigpt},
the conversation template is as follows, 
\begin{center}
    \vspace{-1mm}
    \begin{tabular}{|p{0.95\textwidth}|}
        \hline
        \textit{[INST]~\textless Img\textgreater~\textless Image Feature\textgreater ~ \textless/Img\textgreater~[Task Identifier]~\textless Prompt Feature\textgreater} 
        \textit{[/INST]\qquad} \\
        \hline
    \end{tabular}
    \vspace{-1mm}
\end{center} 
where \textit{[INST]} and \textit{[/INST]}  represent the user role and chat assistant, respectively.
During training, the \textless\textit{Image Feature}\textgreater~is replaced with the visual embeddings, and the \textless\textit{Prompt Feature}\textgreater~is substituted with textual prompt embeddings. 
The \textit{[Task Identifier]} is replaced based on specific circumstances (e.g., \textit{[vqa]} and \textit{[caption]}), making our model more adept at understanding multiple tasks. 

We initialize our model with the pre-trained parameters of MiniGPT-V2~\cite{chen2023minigpt}. MiniGPT-V2 has demonstrated impressive performance across multiple tasks, having undergone three stages of training on various fine-grained datasets~\cite{schuhmann2021laion, sharma2018conceptual2, ordonez2011im2text, lin2014microsoft, sidorov2020textcaps, kazemzadeh2014referitgame, yu2016modeling, mao2016generation, hudson2019gqa, goyal2017making, mishra2019ocr, marino2019ok, schwenk2022okvqa, plummer2015flickr30k}.
We fine-tune the model on downstream tasks at a low cost to validate the effectiveness of our method. Our model is trained on the COCO-QA dataset~\cite{ren2015exploring}, which includes a mix of question-answer pairs covering object classification, color recognition, counting, and positional reasoning. 
The training image-text pairs account for around $67\%$, and the remaining pairs are used for the zero-shot image-to-text generation task.
Throughout the entire training process, the text encoder, global prompt-aware vision encoder, and LLM remain frozen.
Only the local prompt-aware vision adapter and additional projection layers are fine-tuned. 
Evaluation is primarily conducted on the COCO-QA~\cite{ren2015exploring} test dataset and the MME~\cite{fu2023mme} benchmark.
The latter includes $10$ perceptual (i.e., existence, count, position, color, poster, celebrity, scene, landmark, artwork, and OCR) and $4$ cognitive (i.e., commonsense reasoning, numerical calculation, text translation, and code reasoning) tasks.

\begin{figure*}[t!]
\centering
\includegraphics[width=1.\textwidth]{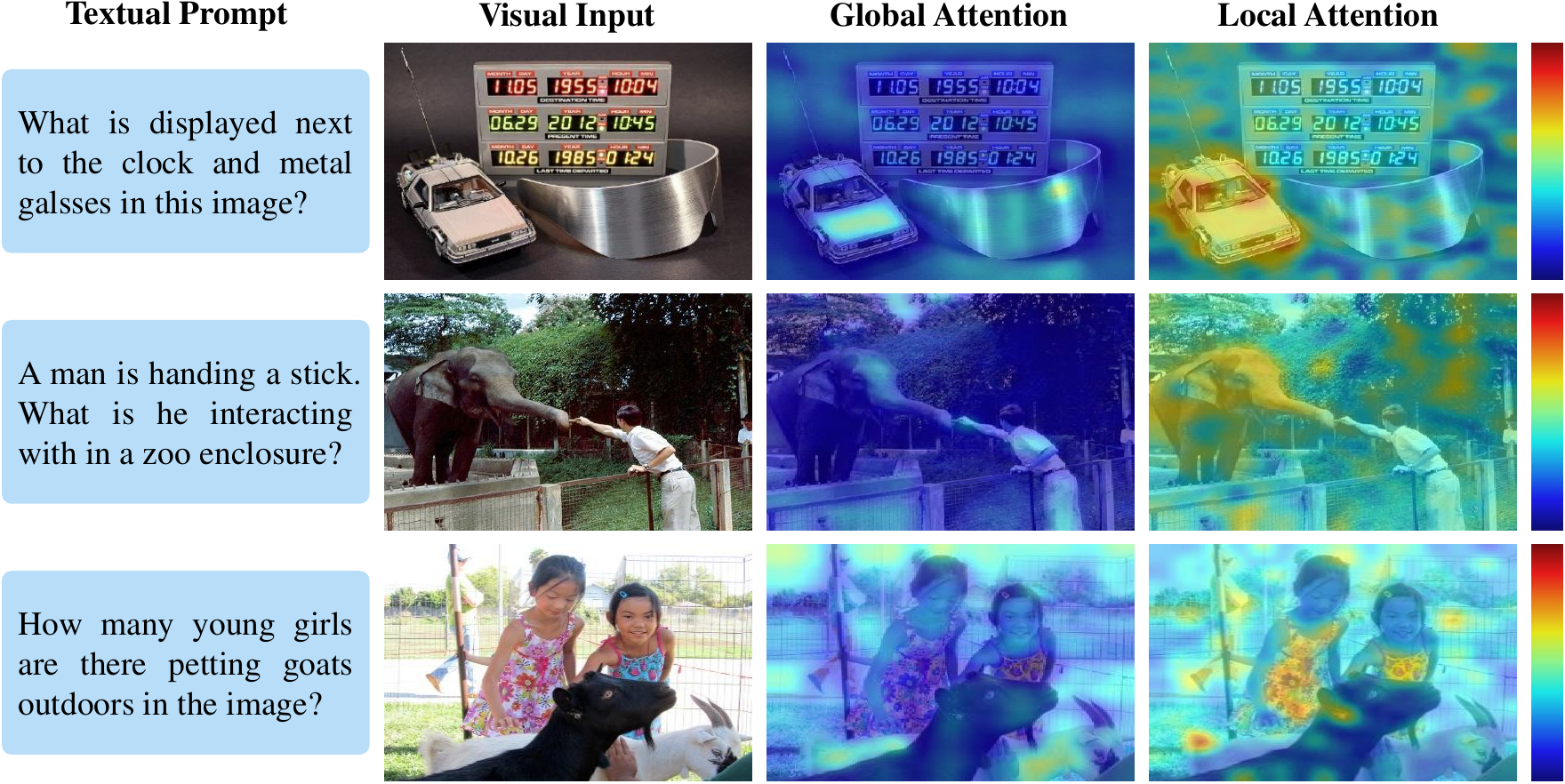}
\vspace{-2em}
\caption{Visualization of prompt-aware global and local attention.
Global attention spans the entire prompt content, while local attention concentrates predominantly on the specific object in question.} 
\vspace{-1em}
\label{fig:attn}
\end{figure*}
\subsection{Evaluation}\label{sec:eval}
\textbf{Attention Visualization.} 
To demonstrate the effectiveness of the proposed method, we visualize  global and local attention in Fig.~\ref{fig:attn}.
The global attention map effectively targets entire regions of the image that semantically align with the entire prompt sentence. The local attention map, associated with the question, highlights its ability to explore fine-grained semantics when extracting visual clues. 

\begin{table*}[t!]
    \centering\setlength{\tabcolsep}{2pt}\small
    \caption{Quantitative results of $10$ perception tasks on the MME~\cite{fu2023mme} benchmark. Evaluation metric is scores for correct answers, with higher scores indicating better performance. $\ast$: frozen, $\checkmark$: fine-tuned. The best and second-best scores are highlighted in bold and underlined, respectively.}
    \resizebox{\textwidth}{!}{
    \begin{tabular}{cccccccccccc}
        \toprule
        Method & Existence & Count & Position & Color & Poster & Celebrity & Scene & Landmark & Artwork & OCR & Total \\
        \midrule
        MiniGPT-4~\cite{zhu2023minigpt} & 68.33 & 55.00 & 43.33 & 75.00 & 41.84 & 54.41 & 71.75 & 54.00 & 60.50 & 57.50 & 581.66 \\ 
        PandaGPT~\cite{su2023pandagpt} & 70.00 &  50.00 & 50.00 & 50.00 & 76.53 & 57.06 & 118.00 & 69.75 & 51.25 &  50.00 & 642.59 \\
        ImageBind-LLM~\cite{zhang2023llama} & 128.33 & 60.00 & 46.67 &  73.33 & 64.97 & 76.47 & 113.25 & 62.00 & 70.75 & 80.00 &  775.77\\          
        VPGTrans~\cite{zhang2024vpgtrans} & 70.00 & 85.00 & 63.33 & 73.33 & 84.01 & 53.53 & 141.75 & 64.75 & 77.25 & 77.50 & 790.45\\   
        LaVIN~\cite{luo2024cheap} & \textbf{185.00} & 88.33 & 63.33 & 75.00 & 79.59 & 47.35 & 136.75 & 93.50 & 87.25 & 107.50 &	963.60\\
        mPLUG-Owl~\cite{ye2023mplug} & 120.00 & 50.00 & 50.00 & 55.00 & 136.05 & 100.29 & 	135.50 & \underline{159.25} & 96.25 & 	65.00 & 967.34 \\
        LRV-Instruction~\cite{liu2023mitigating} & 165.00 & 111.67 & \underline{86.67} & \textbf{165.00} & 139.04 & \underline{112.65} & 147.98 & \textbf{160.53} &  101.25 & \underline{110.00} & \underline{1299.79} \\ 
        BLIP-2~\cite{li2023blip} & 160.00 & \underline{135.00} & 77.33 & 148.33 & \underline{141.84} & 105.59 & 145.25 & 138.00 & \textbf{136.50} & \underline{110.00} & 1293.84 \\
        \midrule   
        Flamingo~\cite{alayrac2022flamingo} & 70.00 & 60.00 & 46.67 & 73.33 & 41.84 & 57.06 & 68.00 & 62.00 & 60.50 & 65.00 & 604.40\\
        Multimodal-GPT~\cite{gong2023multimodal} & 61.67 & 55.00 & 58.33 & 68.33 & 57.82 & 73.82 & 68.00 & 69.75 & 59.50 & 82.50 &  654.72  \\
        InstructBLIP~\cite{dai2024instructblip} & \textbf{185.00} & \textbf{143.33} & 66.67 & \underline{153.33} & 123.81 & 101.18 & \underline{153.00} & 79.75 & \underline{134.25} & 72.50  & 1212.82 \\
        \midrule
         Linear$\checkmark$  & 88.33 & 35.00 & 20.00 & 43.33 & 24.82 & 93.82 & 74.75 & 108.75 & 41.50 & 27.5 & 557.82\\
        Q-Former$\ast$ + Linear$\checkmark$ &  83.18 & 49.45& 50.66 & 68.95 & 55.05 & 55.68 & 49.93 & 53.02 & 30.29 & 49.54 &545.78\\     
        Q-Former$\checkmark$ + Linear$\checkmark$ & 65.00 & 33.22 & 51.66 & 54.21 & 42.70 & 52.60 & 73.75 &  73.04 & 52.13 & 72.50 & 570.84 \\
        Cross-Attention$\checkmark$ & 91.66 & 63.33 & 39.99 & 21.66 & 57.14 & 38.23 & 110.25 & 99.25 & 76.75 & 70.00 & 668.29\\
        \midrule
        \textbf{Ours} & \textbf{185.00} & \underline{135.00} & \textbf{99.99} & 99.99 & \textbf{144.89} & \textbf{141.47} & \textbf{159.00} & 145.15 & 132.00 & \textbf{132.50} & \textbf{1375.02} \\     
        \bottomrule
    \end{tabular}
    }
    \vspace{-5mm}
\label{tab:mme-per}
\end{table*}

\textbf{Quantitative Results.}\label{sec:quanti-eval}
For quantitative evaluation, we exclusively consider questions with precise and concise answers, due to the statistical challenges posed by open-ended responses from MLLMs~\cite{fu2023mme, openai2022chatgpt, liu2023visual}.
First, we compare our model with MLLMs equipped with prompt-unware adapters, such as MiniGPT-4~\cite{zhu2023minigpt}, mPLUG-Owl~\cite{ye2023mplug}, BLIP-2~\cite{li2023blip}, and MLLMs that consider prompts during visual encoding, like Multimodal-GPT~\cite{gong2023multimodal}, InstructBLIP~\cite{dai2024instructblip}.
The above comparison models are zero-shot evaluation results. 
Second, we compare the proposed prompt-aware adapters with other popular adapters. 
To ensure fairness, we strive to exclude the influence of training data and the number of parameters on MLLM performance.
Hence, we use the same dataset~\cite{ren2015exploring} to train MLLMs with MiniGPT-4/v2~\cite{zhu2023minigpt, chen2023minigpt} as the uniform backbone, changing only the adapters, including linear projection, Q-Former family, cross-attention, and our prompt-aware adapter. Results are from the MME leaderboards\footnote{https://github.com/BradyFU/Awesome-Multimodal-Large-Language-Models/tree/Evaluation} 

\begin{wraptable}{r}{0.6\linewidth}
\vspace{-2mm}
    \centering 
    \caption{Quantitative results of object classification, counting, color recognition, and position reasoning on the COCO-QA~\cite{ren2015exploring} test dataset. Evaluation metric is accuracy (\%). $\ast$: frozen parameters, $\checkmark$: fine-tuned parameters. The best results are highlighted in bold.} 
    \resizebox{0.6\textwidth}{!}{
    \begin{tabular}{cccccc}
        \toprule
        Method & Object  & Count  & Color & Position & Total  \\ 
        \midrule
        MiniGPT-4~\cite{zhu2023minigpt} & 73.69 & 36.01 & 63.32 &57.23 & 68.42 \\      
        Linear$\checkmark$ & \underline{74.86} & 62.15& 69.87 & 63.97 & 72.44 \\     
        Q-Former$\ast$ + Linear${\checkmark}$& 56.50 & 43.04 & 29.75 & 35.23 & 49.73\\  
        Q-Former${\checkmark}$ + Linear${\checkmark}$ & 28.46 & 38.83 & 32.09 & 26.43 &  29.67\\
        Cross-Attention${\checkmark}$ & 76.47 & \underline{71.79} & \underline{69.95} & \underline{65.98} & \underline{74.39}   \\   
        \midrule
        \textbf{Ours} & \textbf{81.12} & \textbf{76.19} & \textbf{80.17} & \textbf{70.70} & \textbf{79.95}   \\    
        \bottomrule
    \end{tabular}
    }
    \vspace{-3mm}
\label{tab:coco}
\end{wraptable}
Tab.~\ref{tab:coco} shows the quantitative results on the COCO-QA~\cite{ren2015exploring} text dataset.
The proposed method is capable of selectively focusing on objects mentioned in the question during visual feature extraction, leading to a noticeable improvement (approximately 5.78\%) in quantity perception compared to cross-attention adapters.
In the object classification, color recognition and positional reasoning tasks, our prompt-aware adapters surpass cross-attention adapters by more than 5.73\%, 12.75\%, and 6.95\%, respectively.
As depicted in Tab.s~\ref{tab:mme-per} and~\ref{tab:mme-cog}, compared to prompt-unaware adapters, our method excels in both perception tasks ($1375.02$ \textit{vs} $1299.79$) and cognition tasks ($289.28$ \textit{vs} $210.31$) on the MME~\cite{fu2023mme} benchmark.

\begin{figure}[t!]
\vspace{-3mm}
\centering
\includegraphics[width=\textwidth]{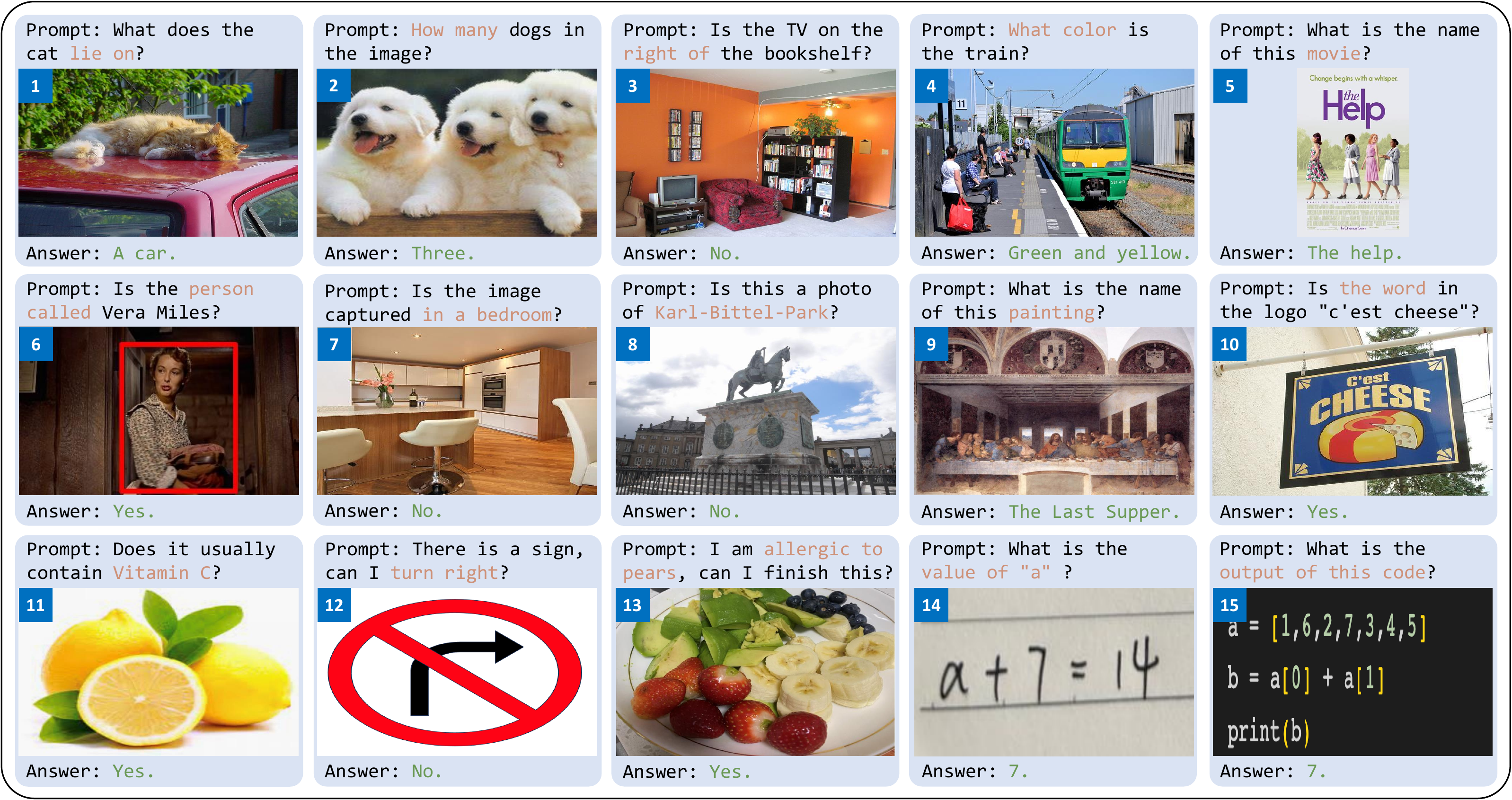}
\vspace{-1em}
\caption{Qualitative results of the proposed method on  diverse perception and cognition tasks.}
\label{fig:visual-per-cog}
\vspace{-5mm}
\end{figure}

\begin{figure*}[t!]
\vspace{-3mm}
\centering
\includegraphics[width=\textwidth]{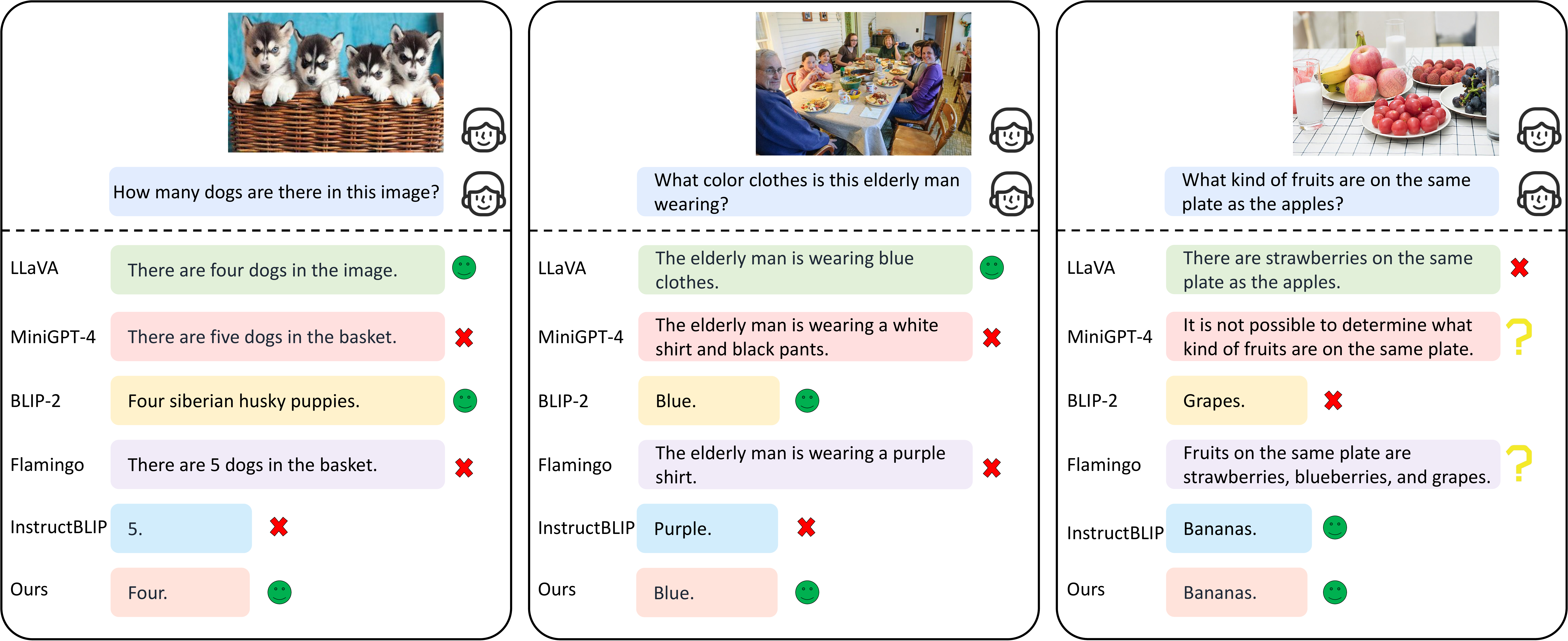}
\caption{Comparison of qualitative results between our method and other popular MLLMs.}
\label{fig:visual-comp}
\vspace{-2mm}
\end{figure*}

\textbf{Qualitative Results.} 
We qualitatively compare our model with several popular MLLMs on perception and cognition tasks with more diverse visual inputs and prompts.
Among the compared methods, LLaVA~\cite{liu2023visual}, MiniGPT-4~\cite{zhu2023minigpt}, BLIP-2~\cite{li2023blip} adopt prompt-unaware adapters.
While Flamingo~\cite{alayrac2022flamingo} and InstructBLIP~\cite{dai2024instructblip} inherently extract prompt-related visual signals through cross-attention.
For visual results, refer to Fig.s~\ref{fig:visual-comp}$\sim$\ref{fig:visual-per-cog} and Appendix.
Thanks to the prompt-aware global and local attention, our model shows a clear improvement in following prompts to focus on specific visual clues.

\subsection{Ablation Study}\label{sec:abla}

\begin{wraptable}{r}{0.55\linewidth}
\vspace{-6mm}
    \centering \setlength{\tabcolsep}{2pt}\small
    \caption{Quantitative results of four cognition tasks on the MME~\cite{fu2023mme} benchmark. Evaluation metric is scores for correct answers, with higher scores indicating better performance. $\ast$: frozen, $\checkmark$: fine-tuned.}
    \resizebox{0.55\textwidth}{!}{
    \begin{tabular}{cccccc}
        \toprule
        Method &  Common &  Calculate & Translate & Code & Total\\
        \midrule
        MiniGPT-4~\cite{zhu2023minigpt} & \underline{59.29} & 45.00 & 0.00 & 40.00 & 144.29 \\
        Linear$\checkmark$ & 58.57 & 42.50 & 5.00 & 47.50 & 153.57 \\
        Q-Former$\ast$ + Linear$\checkmark$ & 37.24 & \textbf{78.07} & \underline{50.00} & 45.00 & \underline{210.31}\\
        Q-Former$\checkmark$ + Linear$\checkmark$ & 38.45 & 0.00 & 47.50 & 5.00 &  90.95\\
        Cross-Attention$\checkmark$ & 40.00  & \underline{52.50} & 15.00 & \textbf{57.50} & 165.00 \\
        \midrule
        \textbf{Ours} & \textbf{99.28} & 50.00 & \textbf{87.50} & \underline{52.50} & \textbf{289.28} \\
        \bottomrule
    \end{tabular}
    }
    \vspace{-5mm}
\label{tab:mme-cog}
\end{wraptable}
\textbf{Ablation Details.} 
In this section, we analyze the necessity of global and local attention. During validation for the former, the global text token is not added to the visual encoder for guidance, while for the latter, a simple linear projection layer is employed to replace the designed local attention component. 
We also test the baseline that does not incorporate any prompt-related attention mechanism during visual feature extraction.
To ensure fairness, the other experimental settings are identical.

\textbf{Prompt-Aware Global Attention.}
Results (see Tab.s~\ref{tab:ab-mme-per}$\sim$\ref{tab:ab-mme-cog}) suggest that the prompt-unaware method struggles to perceive visual inputs and make reasonable inferences, as it does not focus on visual signals emphasized by prompts.
The global attention component directs attention allocation with complete prompt embeddings in the initial phase, thus extracting semantically aligned visual embeddings. 
Without prompt-aware global attention, the answer accuracy of MLLM drops obviously.

\textbf{Prompt-Aware Local Attention.}
We evaluate the local prompt-aware adapter against linear projection without local attention.
According to the results in Tab.s~\ref{tab:ab-mme-per}$\sim$\ref{tab:ab-mme-cog}, the local prompt-aware attention component effectively contributes to the perceptual and cognitive capabilities of our MLLM.
Especially in counting and position reasoning tasks, the correct scores are enhanced by $16.05\%$ and $31.66\%$ (shown in Tab.~\ref{tab:ab-mme-per}), respectively.
Responding to fine-grained prompt guidance during the visual translation process relieves the comprehension load of LLMs.

\begin{table*}[t!]
    \centering\setlength{\tabcolsep}{2pt}\small
    \caption{Ablation study of  prompt-aware global and local attention on MME~\cite{fu2023mme} perception tasks. The best and second-best scores are highlighted in bold and underlined, respectively.}
    \resizebox{\textwidth}{!}{
    \begin{tabular}{cccccccccccc}
        \toprule
        Method & Existence & Count & Position & Color & Posters & Celebrity & Scene & Landmark & Artwork & OCR & Total \\
        \midrule
        prompt-unware & 88.33 & 35.00 & 20.00 & 43.33 & 24.82 & 93.82 & 74.75 & 108.75 & 41.50 & 27.5 & 557.82\\
        \textit{w/o} global-atten & \underline{175.00} & \underline{126.66} & \underline{91.66} & \underline{90.00} & \underline{152.04} & 129.11 & 147.25 & 117.50 & \underline{122.75} & \textbf{132.50} 	 &  \underline{1284.49}\\
        \textit{w/o} local-atten & \underline{175.00} & 113.33  &  68.33 & \textbf{99.99} &  \textbf{152.38} & \underline{131.17} &  \underline{155.25} & \underline{119.75} & \underline{122.75} &  \underline{80.00} & 1217.97\\
        \midrule
        \textbf{w/ global + local} & \textbf{185.00} & \textbf{135.00} & \textbf{99.99} & \textbf{99.99} & 144.89 & \textbf{141.47} & \textbf{159.00} & \textbf{145.15} & \textbf{132.00} & \textbf{132.50} & \textbf{1375.02} \\
        \bottomrule
    \end{tabular}
    }
\label{tab:ab-mme-per}
\end{table*}

\begin{table}[t!]
    \centering
        \begin{minipage}{0.48\textwidth}
        \centering \setlength{\tabcolsep}{1.5pt}\small
        \caption{Ablations on COCO-QA~\cite{ren2015exploring} test set. }
        \vspace{-2mm}
    \begin{tabular}{cccccc}
        \toprule 
        Method & Object  & Count  & Color & Position & Total  \\ 
        \midrule   
         prompt-unware & 74.86 & 62.15 & 69.87 & 63.97 & 72.44 \\
         \textit{w/o} global-atten &  \underline{78.67} & 73.94 &  \underline{79.12} & \textbf{68.40}  & \underline{77.75}\\         
         \textit{w/o} local-atten & 77.08 & \underline{74.59} & 76.06 &  67.83 & 76.15\\
         \midrule
        \textbf{w/ global + local} & \textbf{81.12} & \textbf{76.19} & \textbf{80.17} & \textbf{70.70} & \textbf{79.95}   \\    
        \bottomrule
    \end{tabular}
    \label{tab:ab-coco-per}
    \end{minipage}
    \vspace{-2mm}
    \hfill
    \begin{minipage}{0.48\textwidth}
        \centering \setlength{\tabcolsep}{1.5pt}\small
     \caption{Ablations on MME~\cite{fu2023mme} cognition tasks. }
     \vspace{-2mm}
    \begin{tabular}{cccccc}
        \toprule
        Method &  Comm. &  Cal. & Trans. & Code & Total\\
        \midrule
       prompt-unware & 58.57 & 42.50 & 5.00 & 47.50 & 153.57 \\
        \textit{w/o} global-atten & \underline{96.42} & \textbf{50.00} & \underline{55.00} & \textbf{55.00} & \underline{256.42} \\       
       \textit{w/o} local-atten & \textbf{99.28} & \underline{47.50} & 50.00 & \underline{52.50} & 249.28 \\
        \midrule
        \textbf{w/ global + local} & \textbf{99.28} & \textbf{50.00} & \textbf{87.50} & \underline{52.50} & \textbf{289.28} \\
        \bottomrule
    \end{tabular}
        \label{tab:ab-mme-cog}
    \end{minipage}
    \vspace{-2mm}
\end{table}

\section{Conclusions}
In this paper, we introduce novel prompt-aware adapters designed to adaptively embed visual inputs based on the given prompt. Our method aims to extract the most informative visual clues relevant to the prompt, thereby enhancing the visual understanding capabilities of LLMs. We first present a global attention mechanism that utilizes a global text token for coarse-grained visual guidance.
Then, we introduce a local attention mechanism that leverages local textual features to extract highly relevant visual clues with fine granularity. The resulting prompt-aware global and local visual tokens significantly reduce the visual perception burden on LLMs. Experimental results show that our method achieves competitive accuracy on multiple visual perception and cognitive tasks. We discuss the limitations and broader impacts of our method in Appendix.

\small
\bibliographystyle{unsrt}
\bibliography{neurips_2024}

\newpage
\appendix
\section{Appendix Overview}
In summary, the appendix is organized as follows:

Appendix~\ref{sec:app-exper}: Additional experiments and analysis.

Appendix~\ref{sec:app-attn}: More visualization results of prompt-aware global and local attention.

Appendix~\ref{sec:app-qual}: More qualitative results and analysis.




Appendix~\ref{sec:app-data}: More details of COCO-QA dataset and MME benchmark.

Appendix~\ref{sec:app-prompt}: Prompts used for fine-tuning.



\section{Additional Experiments and Analysis}\label{sec:app-exper}
\subsection{Comparison of Text Encoders}
During the implementation of the global and local attention mechanisms, we compare the quantitative results of text feature extraction by different text encoders.
Tab.s~\ref{tab:mme-per-text}$\sim$\ref{tab:coco-per-text} show comparison results of adopting CLIP~\cite{radford2021learning} and Llama 2~\cite{touvron2023llama} to extract prompt features for global and local attention.
G-Num refers to the number of text tokens used in global attention.
$0$ means no global attention is used in the model architecture, $1$ means only one global text token is used, and $L$ means the total number of text tokens.

The experimental results show that using global text tokens performs better than not using them, confirming the effectiveness of our prompt-aware global attention. Besides, CLIP~\cite{radford2021learning} generally performs better than Llama 2~\cite{touvron2023llama} on perceptual tasks, while Llama 2~\cite{touvron2023llama} excels in cognitive tasks. 
This may be because, in perceptual tasks, the text features extracted by CLIP can better align with visual features. In cognitive tasks, Llama 2~\cite{touvron2023llama}, having been trained on vast amounts of data, possesses superior reasoning abilities.

\begin{table*}[h]
    \centering\setlength{\tabcolsep}{2pt}\small
    \caption{Comparison results of using different text encoders (CLIP~\cite{radford2021learning} and Llama 2~\cite{touvron2023llama}) to extract prompt features for global and local attention. G-Num refers to the number of text tokens used in global attention, and $L$ is the total amount of text tokens. The evaluation is built on the MME~\cite{fu2023mme} perception tasks, the evaluation metric is correct answer scores, with higher scores being better. The best and second-best scores are highlighted in bold and underlined, respectively.}
    \resizebox{\textwidth}{!}{
    \begin{tabular}{ccc|ccccccccccc}
        \toprule
        \multicolumn{3}{c|}{Method} & \multicolumn{10}{c}{Perception Tasks}  \\
        \midrule
        G-Num & Global & Local  & Existence & Count & Position & Color & Poster & Celebrity & Scene & Landmark & Artwork & OCR & Total\\
        \midrule
        $0$ & \text{-} & \text{-} & 88.33 & 35.00 & 20.00 & 43.33 & 24.82 & 93.82 & 74.75 & 108.75 & 41.50 & 27.5 & 557.82\\
        $0$ & \text{-} &  CLIP & \underline{175.00} & \underline{126.66} & \underline{91.66} & 90.00 & \underline{152.04} & 129.11 & 147.25 & 117.50 & \underline{122.75} & \textbf{132.50} &  \underline{1284.49} \\  
        $1$ & CLIP & \text{-} & \underline{175.00} & 113.33  &  68.33 & \textbf{99.99} &  \textbf{152.38} & 131.17 &  155.25 & 119.75 & \underline{122.75} &  80.00 & 1217.97\\ 
        $1$ & CLIP  & Llama~2 &  170.00 & 121.66 & 48.33 & \textbf{99.99} & 130.27 & 127.94 & 149.50 &  134.25 & 119.50 & \underline{125.00} & 1226.46\\
        $L$ & CLIP & Llama~2 & 170.00 & 106.66 & 48.33 & 70.00 &  128.23 &  118.23 & 154.25 & 139.50 & 119.25 & 102.50  & 1156.96 \\ 
        $L$ & CLIP & CLIP & 160.0 & 111.66 & 73.33 & 90.00 & 133.33 & \underline{131.76} & \underline{158.00} & \textbf{146.50} & 119.25 &  95.00 & 1218.84\\   
        \midrule
        $1$ & CLIP & CLIP & \textbf{185.00} & \textbf{135.00} & \textbf{99.99} & \textbf{99.99} & 144.89 & \textbf{141.47} & \textbf{159.00} & \underline{145.15} & \textbf{132.00} & \textbf{132.50} & \textbf{1375.02} \\          
        \bottomrule
    \end{tabular}
    }
\label{tab:mme-per-text}
\end{table*}

\begin{table*}[h]
    \centering \small
    \caption{Comparison results of using different text encoders (CLIP~\cite{radford2021learning} and Llama 2~\cite{touvron2023llama}) to extract prompt features for global and local attention. G-Num refers to the number of text tokens used in global attention, and $L$ is the total amount of text tokens. The evaluation is built on the MME~\cite{fu2023mme} cognition tasks, the evaluation metric is correct answer scores, with higher scores being better. The best and second-best scores are highlighted in bold and underlined, respectively.}
    \resizebox{0.75\textwidth}{!}{
    \begin{tabular}{ccc|ccccc}
        \toprule
        \multicolumn{3}{c|}{Method} & \multicolumn{5}{c}{Cognition Tasks}  \\
        \midrule
        G-Num & Global & Local  & Common Sense & Calculation & Translation & Code & Total\\
        \midrule
        $0$ & \text{-} & \text{-} & 58.57 & 42.50 & 5.00 & 47.50 & 153.57 \\
        $0$ & \text{-} &  CLIP & \underline{96.42} & \underline{50.00} & 55.00 & \textbf{55.00} & \underline{256.42} \\     
        $1$ & CLIP & \text{-} & \textbf{99.28} & 47.50 & 50.00 & \underline{52.50} & 249.28 \\
        $1$ & CLIP  & Llama~2 & 83.57 & \textbf{57.50} & 57.50 & 47.50 & 246.07\\
        $L$ & CLIP & Llama~2 &  92.14 & \underline{50.00} & \underline{65.00} & 47.50 & 254.64\\ 
        $L$ & CLIP & CLIP & \underline{96.42} & 45.00 & 57.50 & 47.50 & 246.42\\  
        \midrule
        $1$ & CLIP & CLIP &  \textbf{99.28} & \underline{50.00} & \textbf{87.50} & \underline{52.50} & \textbf{289.28} \\          
        \bottomrule
    \end{tabular}
    }
\label{tab:mme-cog-text}
\end{table*}

\begin{table*}[h]
    \centering \small
    \caption{Comparison results of using different text encoders (CLIP~\cite{radford2021learning} and Llama 2~\cite{touvron2023llama}) to extract prompt features for global and local attention. G-Num refers to the number of text tokens used in global attention, and $L$ is the total amount of text tokens. The evaluation is built on the COCO-QA~\cite{ren2015exploring} cognition tasks, the evaluation metric is correct answer scores, with higher scores being better. The best and second-best scores are highlighted in bold and underlined, respectively.}
    \resizebox{0.75\textwidth}{!}{
    \begin{tabular}{ccc|ccccc}
        \toprule
        \multicolumn{3}{c|}{Method} & \multicolumn{5}{c}{Perception Tasks}  \\
        \midrule
        G-Num & Global & Local  & Object & Count & Color & Position & Total\\
        \midrule
        $0$ & \text{-} & \text{-} & 74.86 & 62.15 & 69.87 & 63.97 & 72.44 \\
        $0$ & \text{-} &  CLIP & 78.67 & 73.94 &  \underline{79.12} & 68.40  & 77.75\\     
        $1$ & CLIP & \text{-} &   77.08 & 74.59 & 76.06 &  67.83 & 76.15\\
        $1$ & CLIP  & Llama~2 &  \underline{80.57} & 76.13 &  78.64 & 68.76  & 79.25  \\
        $L$ & CLIP & Llama~2 &   80.41 &  \textbf{76.62} &  77.41 &  \textbf{70.94}   &  79.07\\ 
        $L$ & CLIP & CLIP &    80.51 &  \underline{76.40} &   79.07 &   69.00     & 79.25 \\  
        \midrule
        $1$ & CLIP & CLIP &  \textbf{81.12} & 76.19 & \textbf{80.17} & \underline{70.70} & \textbf{79.95}   \\          
        \bottomrule
    \end{tabular}
    }
\label{tab:coco-per-text}
\end{table*}

\subsection{Residual Visual Details}
We also try to fuse the visual features obtained through global attention and the visual features obtained through local attention.
$Ratio$ represents the retention ratio of local visual features, and the retention ratio of global visual features is (1 - $Ratio$).
For example, when $Ratio$=0.2, it means that the local features are multiplied by 0.2 and the global features are multiplied by 0.8 and then added.

Tab.s ~\ref{tab:app-mme-per-res}$\sim$\ref{tab:app-mme-cog-res} show the impact of using different $Ratio$ values on the model's perceptual and cognitive tasks.
Experimental results show that the weighted sum of global and local features helps to improve model performance to a certain extent but not very much.
In addition, when local features account for greater weight, the model seems to perform better.

\subsection{Exploration of Prompt-Aware Q-Former}
Inspired by InstructBLIP~\cite{li2023blip}, we further explore novel architectures for prompt-aware Q-Former.
In this section, we implement a Q-Former-based prompt-aware adapter.
Specifically, we try to add prompt cross-attention to different blocks in Q-Former.
Table~\ref{tab:app-acc} shows the quantitative evaluation results of various model-specific implementations.
Through extensive experiments, we found that the Q-former is difficult to train, and it tends to lose previously learned knowledge when fine-tuning on downstream tasks. Additionally, the effects of injecting text features into different layers of the Q-Former vary significantly.


\begin{table*}[t!]
    \centering \small
    \caption{Comparison results of using different $Ratio$ to combine global and local visual features. The value range of $Ratio$ is 0-1. The evaluation is built on the MME~\cite{fu2023mme} perception tasks, the evaluation metric is correct answer scores, with higher scores being better. The best and second-best scores are highlighted in bold and underlined, respectively. }
    \resizebox{\textwidth}{!}{
    \begin{tabular}{cccccccccccc}
        \toprule
        $Ratio$ & Existence & Count & Position & Color & Posters & Celebrity & Scene & Landmark & Artwork & OCR & Total \\
        \midrule
      0.2 & 85.00 & 65.00 & 58.33 & 51.66 & 71.42 & 40.29 & 52.25 & 99.75 & 56.25 & 70.00 & 649.97 \\
      0.4 & 175.00 & 108.33 & 68.33 & 73.33 & 130.27 & 136.17 & \underline{156.75} & \textbf{148.25} & \underline{136.00} & \underline{95.00} & 1227.44 \\
      0.6 & 175.00 & 126.66 & \underline{91.66} & 90.00 & \underline{152.04} & 129.11  & 147.25 & 117.50 & 122.75 & \textbf{132.50} & 1284.49 \\ 
      0.8 & \underline{185.00} & \textbf{135.00} & \textbf{99.99} & \underline{99.99} & 144.89 & 141.47 & \textbf{159.00} & \underline{145.15} & 132.00 & \textbf{132.50} & \textbf{1375.02} \\ 
      1.0 & \textbf{190.0} & \underline{128.33}  & \underline{91.66} & \textbf{150.00} & \textbf{152.72} & \textbf{144.41} & 155.50 & 141.00 & \textbf{139.75} & 80.00 & \underline{1373.38}\\    
        \bottomrule
    \end{tabular}
    }
\label{tab:app-mme-per-res}
\end{table*}

\begin{table*}[t!]
\small
    \centering\setlength{\tabcolsep}{2pt}\small
    \caption{Comparison results of using different $Ratio$ to combine global and local visual features. The value range of $Ratio$ is 0-1. The evaluation is built on the MME~\cite{fu2023mme} cognition tasks. the evaluation metric is correct answer scores, with higher scores being better. The best and second-best scores are highlighted in bold and underlined, respectively.}
    \resizebox{0.6\textwidth}{!}{
    \begin{tabular}{cccccc}
        \toprule
        $Ratio$ &  Commonsense &  Calculation & Translation & Code & Total\\
        \midrule
        0.2 & 67.14 & \textbf{65.00} & 5.00 & \underline{47.50} & 184.64 \\
        0.4 & \underline{97.14} & 47.50 & 50.00 & \underline{47.50} & 242.14 \\
        0.6 & 96.42 & 50.00 & 55.00 & \textbf{55.00} & \underline{256.42} \\
        0.8 & 92.85 & \underline{57.50} & \underline{57.50} & 45.00 & 252.85 \\
        1.0 & \textbf{103.57} & 55.00 & \textbf{72.50} & \underline{47.50} & \textbf{278.57}\\ 
        \bottomrule
    \end{tabular}
    }
\label{tab:app-mme-cog-res}
\end{table*}

\begin{table*}[h]
    \centering\setlength{\tabcolsep}{6pt}
    \caption{Quantitative evaluation results of various model-specific implementations. The ``No." is the number of the experimental setup. The ``BlockNum" denotes the number of attention blocks in which the text cross-attention layer is inserted. The ``OC", ``CON", ``CR", and ``PR" represent the four tasks of object classification, counting, color recognition, and position reasoning, respectively. The ``Acc (\%)“ denotes accuracy and ``Total Acc (\%)" refers to the overall accuracy on four tasks.} 
    \resizebox{\textwidth}{!}{
    \begin{tabular}{ccccccc}
        \toprule
        \multirow{2}*{No.} & \multirow{2}*{BlockNum} & OC  & CON  & CR & PR & Total  \\ 
        & &Acc~(\%) & Acc~(\%) & Acc~(\%) & Acc~(\%) & Acc~(\%)\\ 
        \midrule
        1& 0& 78.09 & \textbf{73.22} & 74.23 & 65.54 & 76.33\\
        2& 0+1& 65.46 & 70.84 & 51.55 & 52.31 & 62.73 \\
        3& 0+1+2& 77.67 & 65.48 & 73.49 & 63.08 & 75.23\\  
        4& 0+1+2+3& 78.20 & 69.95 & 69.89 & 62.47 & 75.29 \\
        5& 0+1+2+3+4& 77.72 & 69.35 & 79.06 & 62.47 & 76.38\\
        6& 0+1+2+3+4+5& 78.52 & 71.43 & 74.85 & 67.39 & 76.73\\
        7& 1& 29.31 & 36.31 & 8.68 & 10.47 & 25.23\\
        8& 1+3+5+7+11& 33.21 & 29.47 & 9.79 & 22.77 & 28.50\\
        9& 0+2+4+6+8+10& 77.98 & 69.05 & 72.37 & 64.31 & 75.59\\
        10& 11& 77.02 & 68.46 & 60.97 & 62.77 & 72.93\\
        11& 10+11& 75.15 & 71.73 & 72.37 & 64.62 & 73.79\\
        12& 9+10+11& 74.89 & 69.95 & 72.87 & 66.47 & 73.69\\
        13& 8+9+10+11& 75.09 & 69.95 & 73.49 & 65.85 & 73.89\\
        14& \textbf{7+8+9+10+11}& \textbf{78.65} & 72.03 & \textbf{78.31} & \textbf{67.69} & \textbf{77.26}\\
        15& 6+7+8+9+10+11& 77.72 & 68.76 & 75.97 & 64.92 & 76.1\\
        16& 0+1+2+3+4+5+6+7+8+9+10+11& 76.70 & 69.05 & 74.73 & 64.62 & 75.09\\
        \bottomrule
    \end{tabular}
    }
\label{tab:app-acc}
\end{table*}

\newpage
\section{More Visualization Results of Prompt-Aware Global and Local Attention}\label{sec:app-attn}
\begin{figure}[h]
\centering
\includegraphics[width=\textwidth]{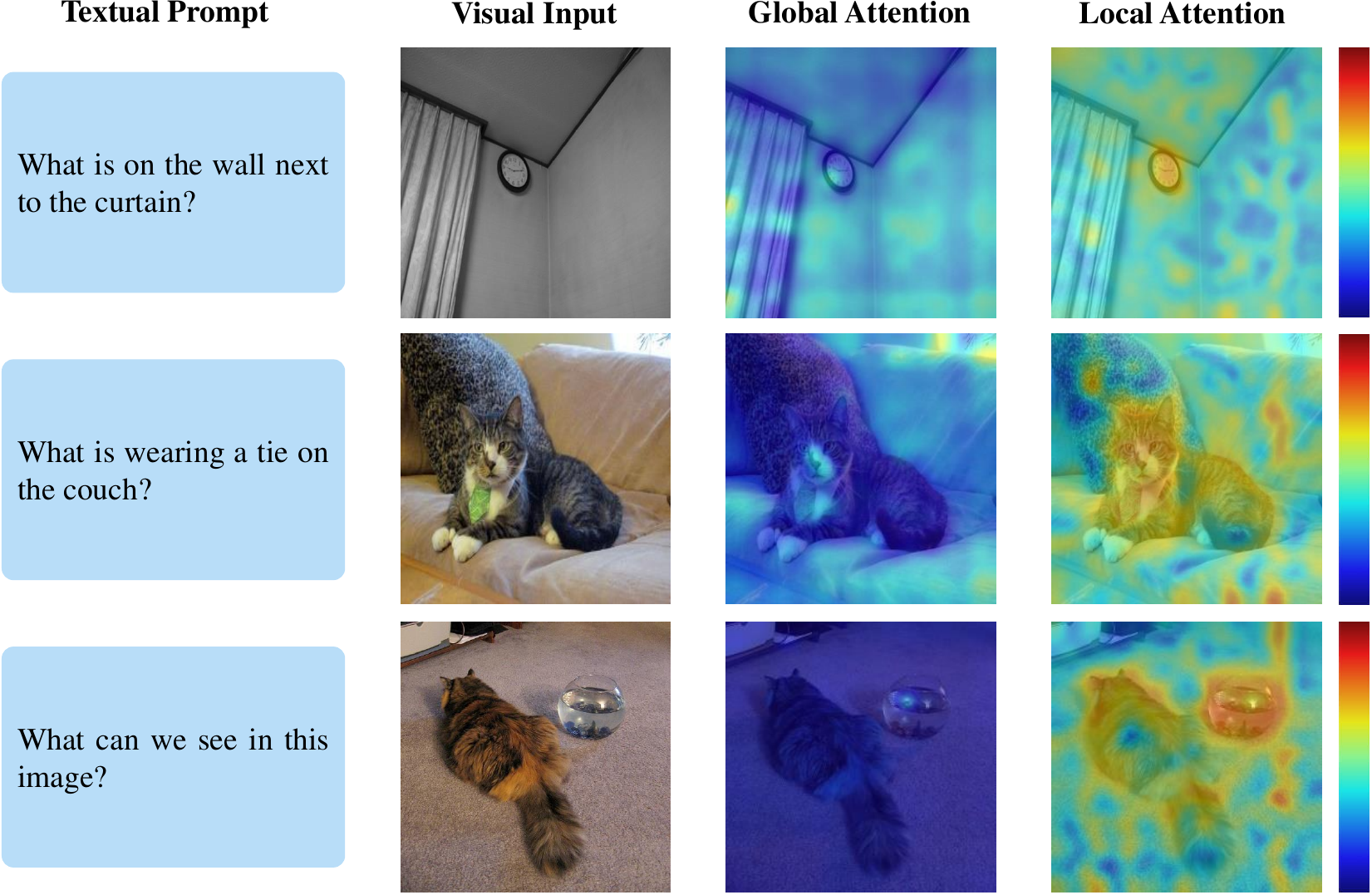}
\caption{Visualization of prompt-aware global and local attention on the object perception task.  Visualization results show that global attention spans the entire prompt content, while local attention concentrates predominantly on the specific object in question.}
\label{fig:app-attn-object}
\end{figure}

\begin{figure}[h]
\centering
\includegraphics[width=\textwidth]{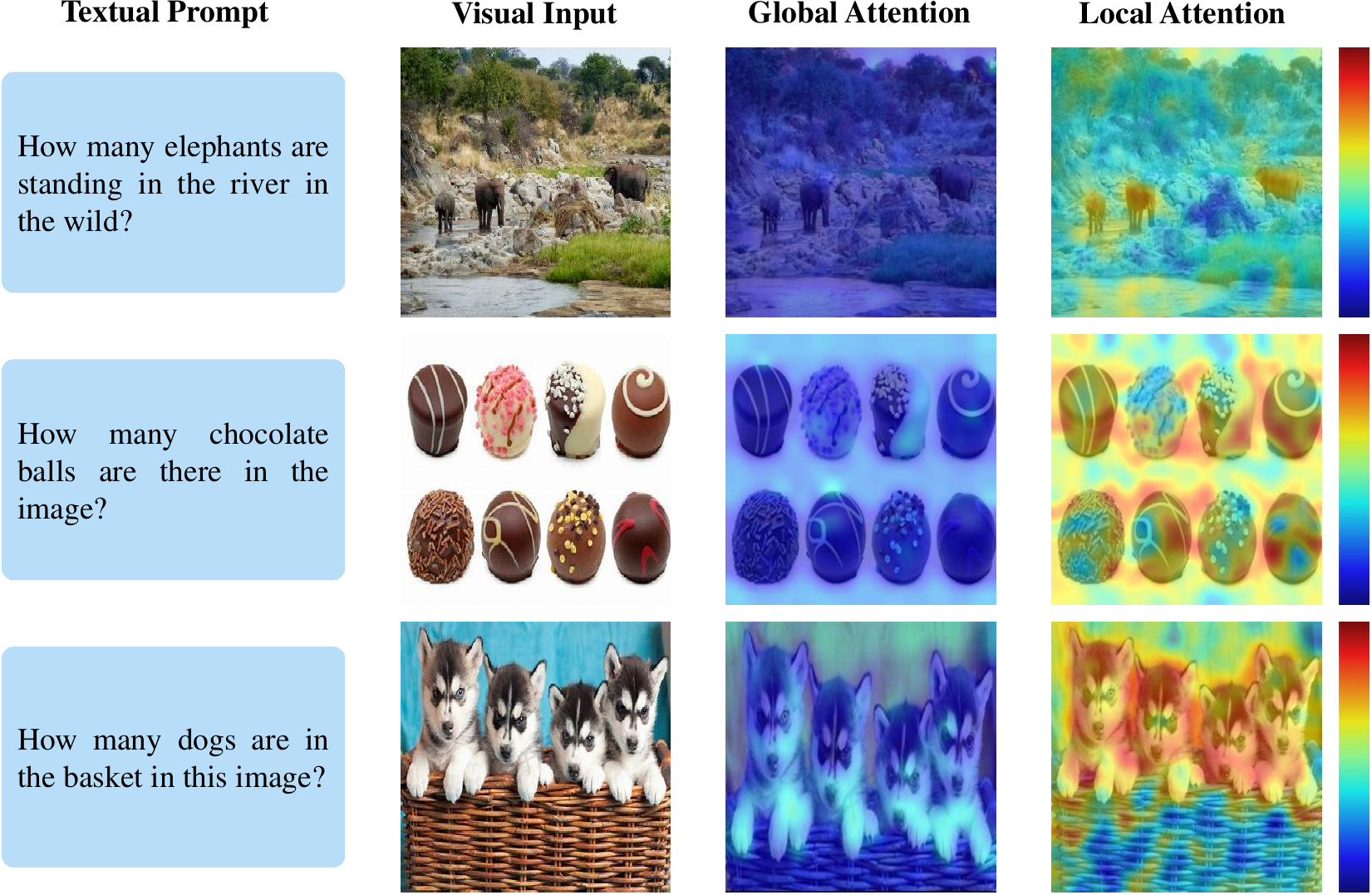}
\caption{Visualization of prompt-aware global and local attention on counting task. Visualization results show that global attention spans the entire prompt content, while local attention concentrates predominantly on the specific object in question. Our method performs well in accurately identifying and counting objects.}
\label{fig:app-attn-count-1}
\end{figure}

\begin{figure}[h]
\centering
\includegraphics[width=\textwidth]{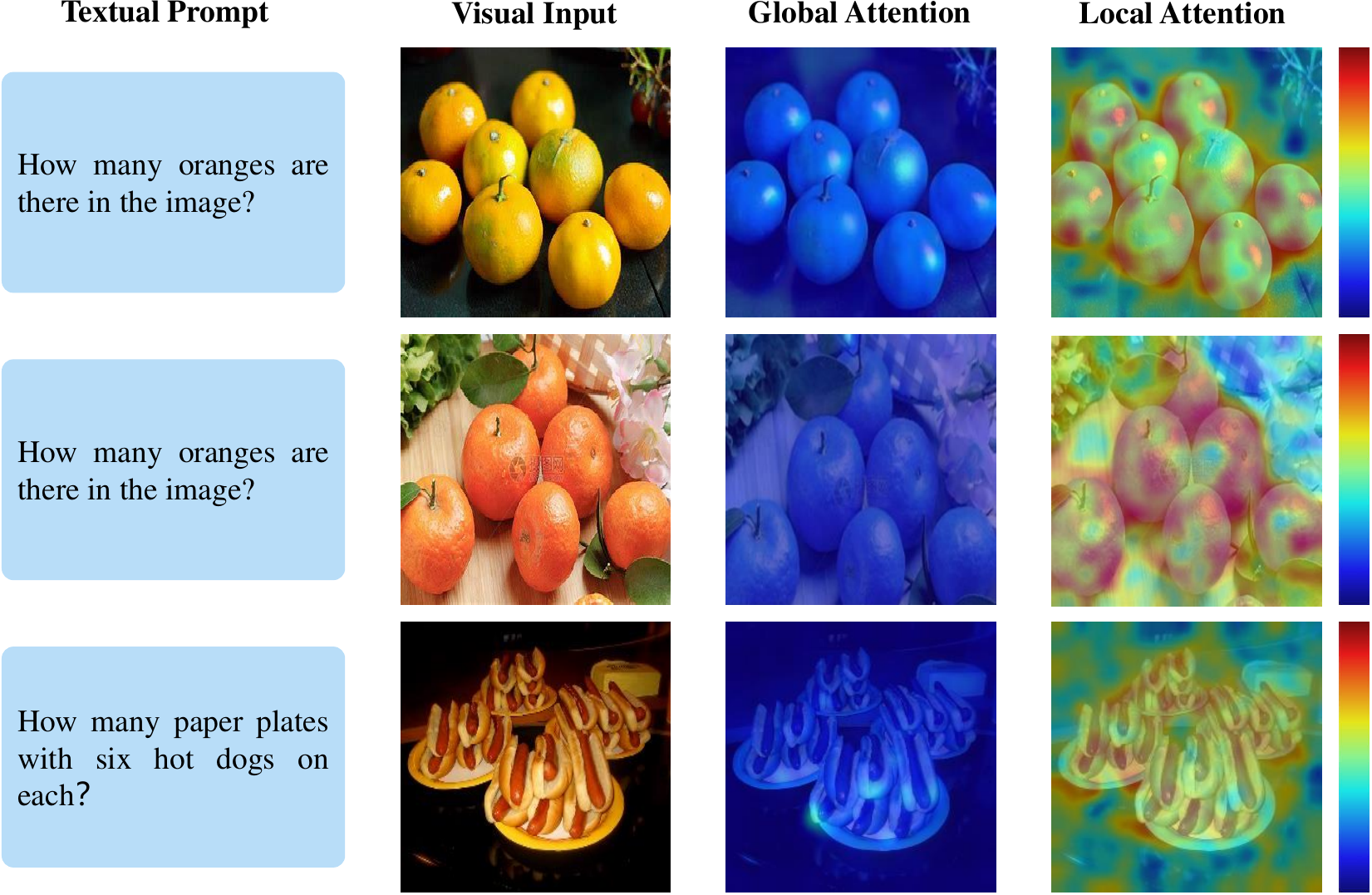}
\caption{Visualization of prompt-aware global and local attention on counting task. Visualization results show that global attention spans the entire prompt content, while local attention concentrates predominantly on the specific object in question. Our method performs well in accurately identifying and counting objects.}
\label{fig:app-attn-count-2}
\end{figure}

\begin{figure}[h]
\centering
\includegraphics[width=\textwidth]{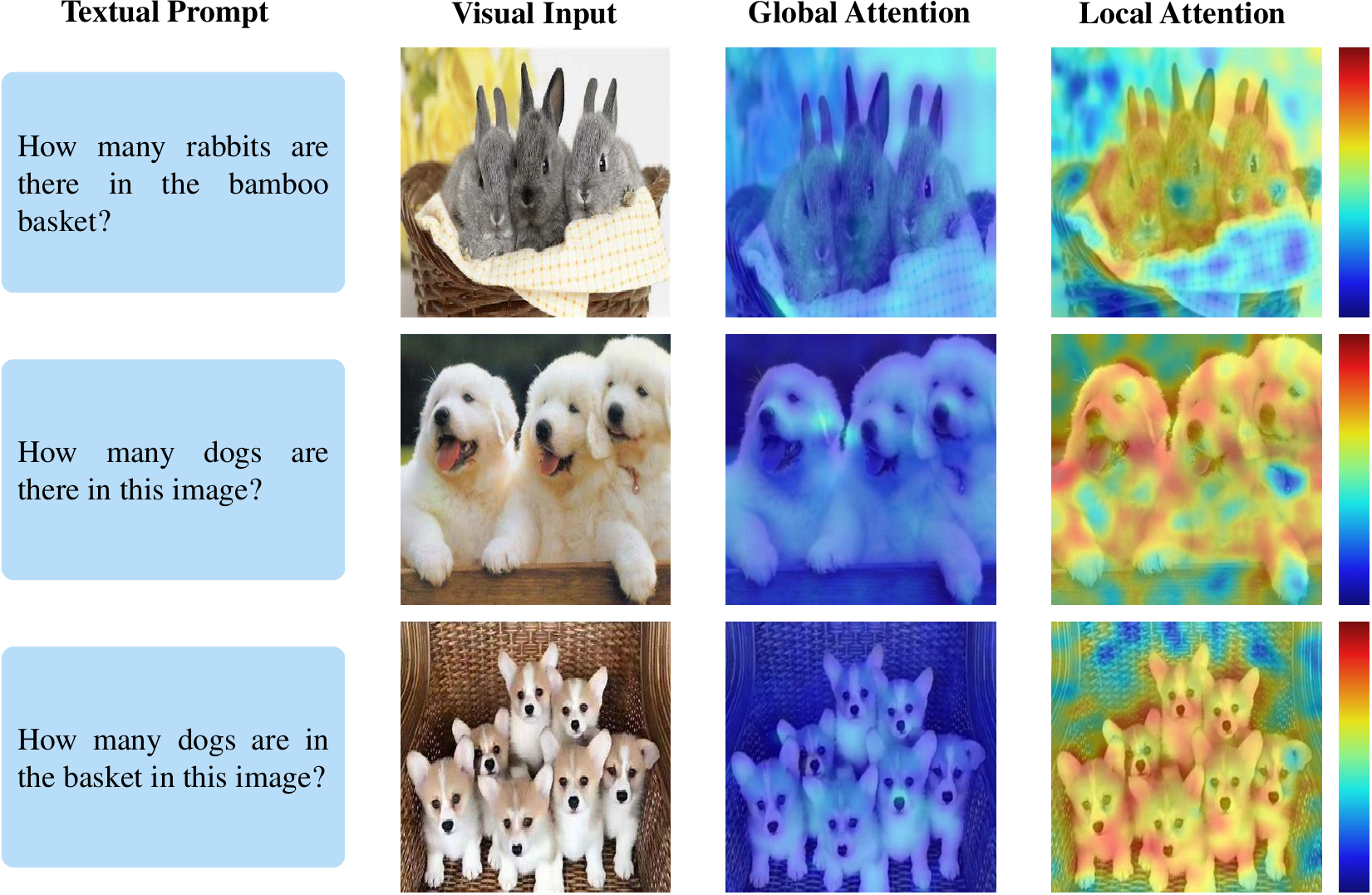}
\caption{Visualization of prompt-aware global and local attention on counting task. Visualization results show that global attention spans the entire prompt content, while local attention concentrates predominantly on the specific object in question. Our method performs well in accurately identifying and counting objects.}
\label{fig:app-attn-count-3}
\end{figure}

\begin{figure}[h]
\centering
\includegraphics[width=\textwidth]{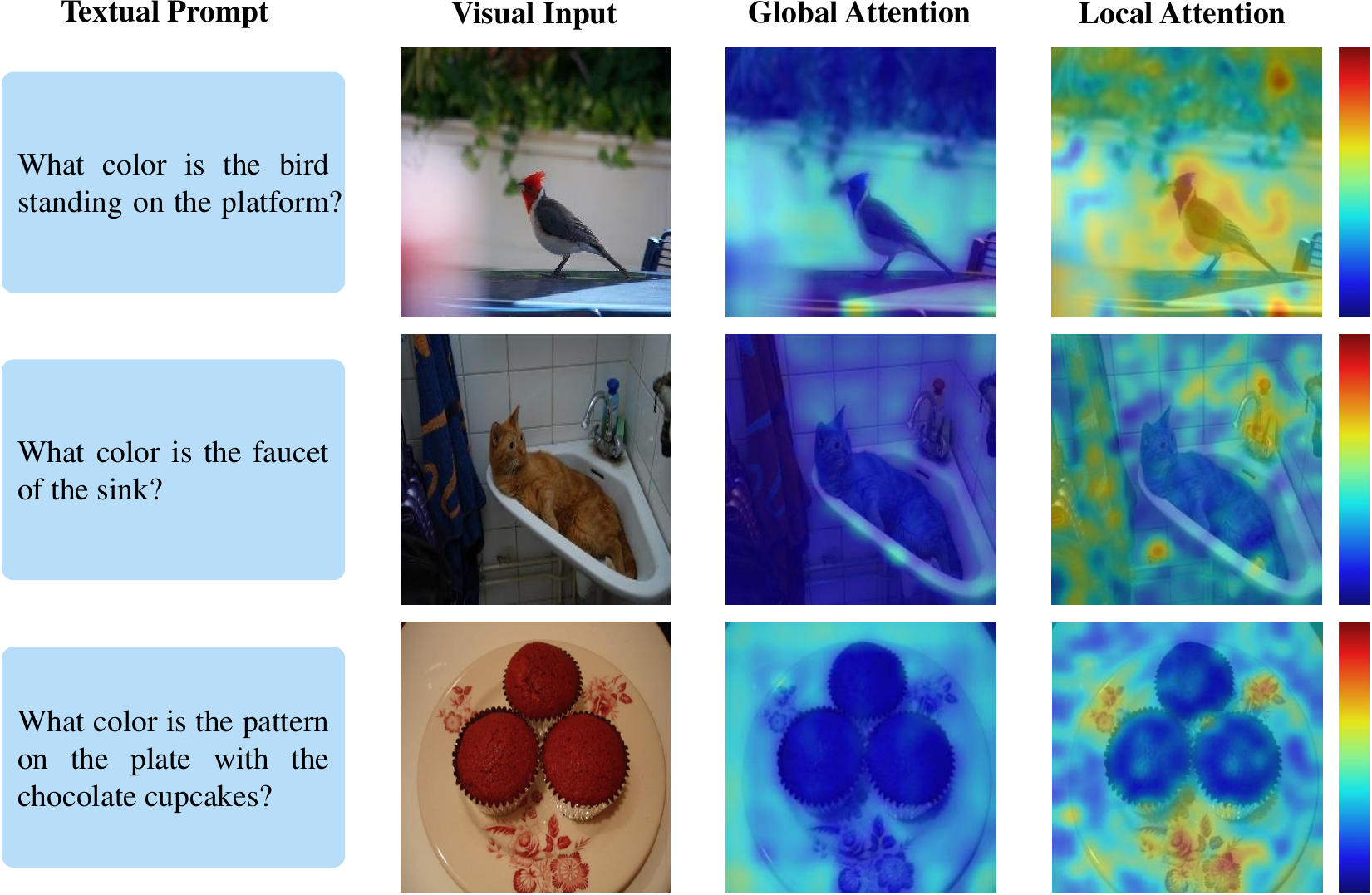}
\caption{Visualization of prompt-aware global and local attention on the color recognition task. Visualization results show that global attention spans the entire prompt content, while local attention concentrates predominantly on the specific object in question. Our method achieves impressive performance in accurately identifying and distinguishing colors.}
\label{fig:app-attn-color}
\end{figure}

\begin{figure}[h]
\centering
\includegraphics[width=\textwidth]{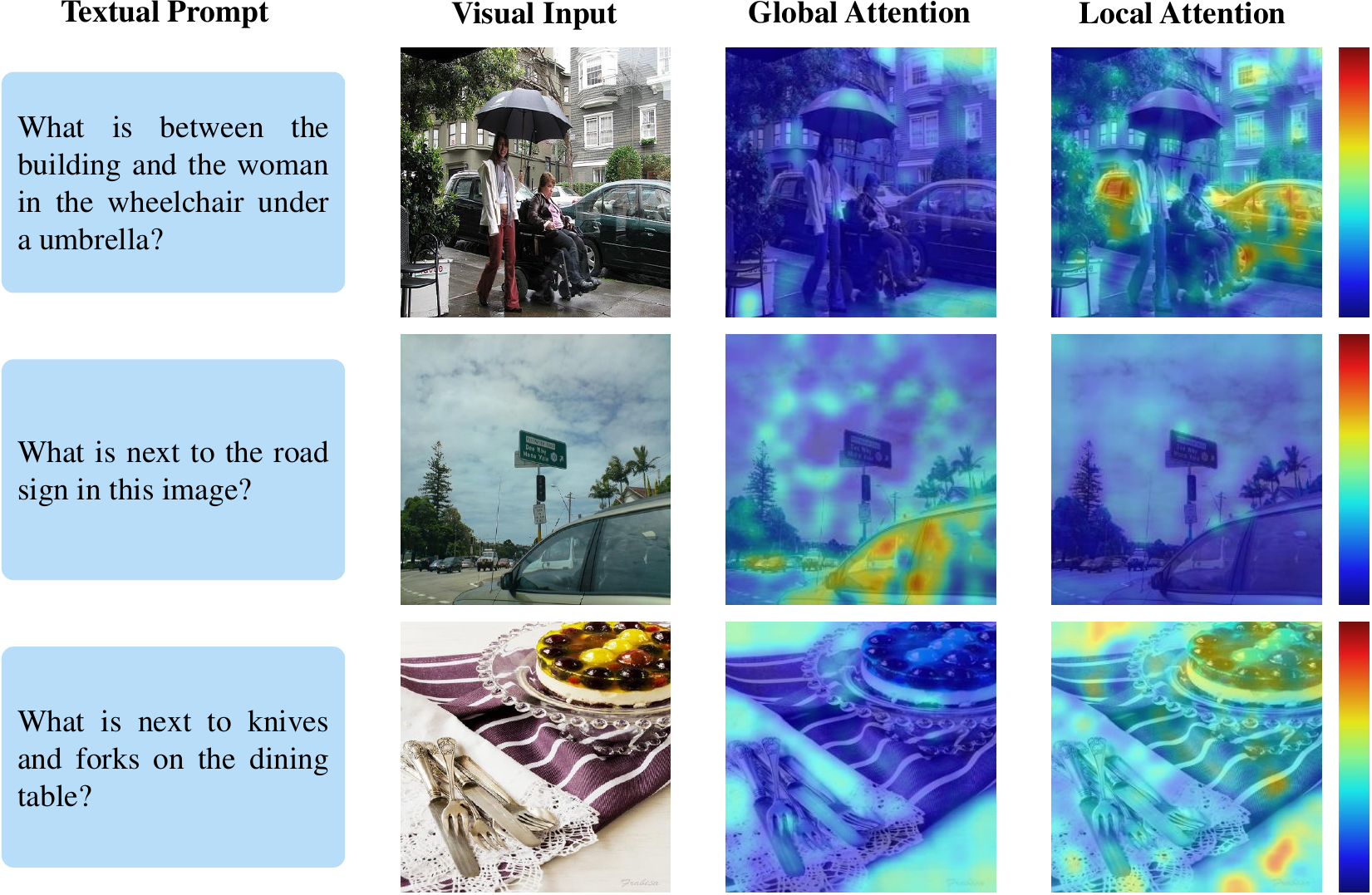}
\caption{Visualization of prompt-aware global and local attention on the position reasoning task. Visualization results show that global attention spans the entire prompt content, while local attention concentrates predominantly on the specific object in question. Our method has the capability of location recognition.}
\label{fig:app-attn-position}
\end{figure}

In this section, we present more results of global and local attention visualization. Fig.~\ref{fig:app-attn-object} aims to verify the role of the proposed attention mechanism in object recognition tasks. Fig.s~\ref{fig:app-attn-count-1} to \ref{fig:app-attn-count-3} re-validate the superiority of the proposed prompt-aware adapter in counting. Fig.~\ref{fig:app-attn-color} and Fig.~\ref{fig:app-attn-position} illustrate the focal points of the attention mechanism in color recognition and position reasoning tasks, respectively. A large number of visualization results illustrate the effectiveness of our method and provide a degree of interpretability.

\section{More Qualitative Results and Analysis}\label{sec:app-qual}
\begin{figure}[h]
\centering
\includegraphics[width=\textwidth]{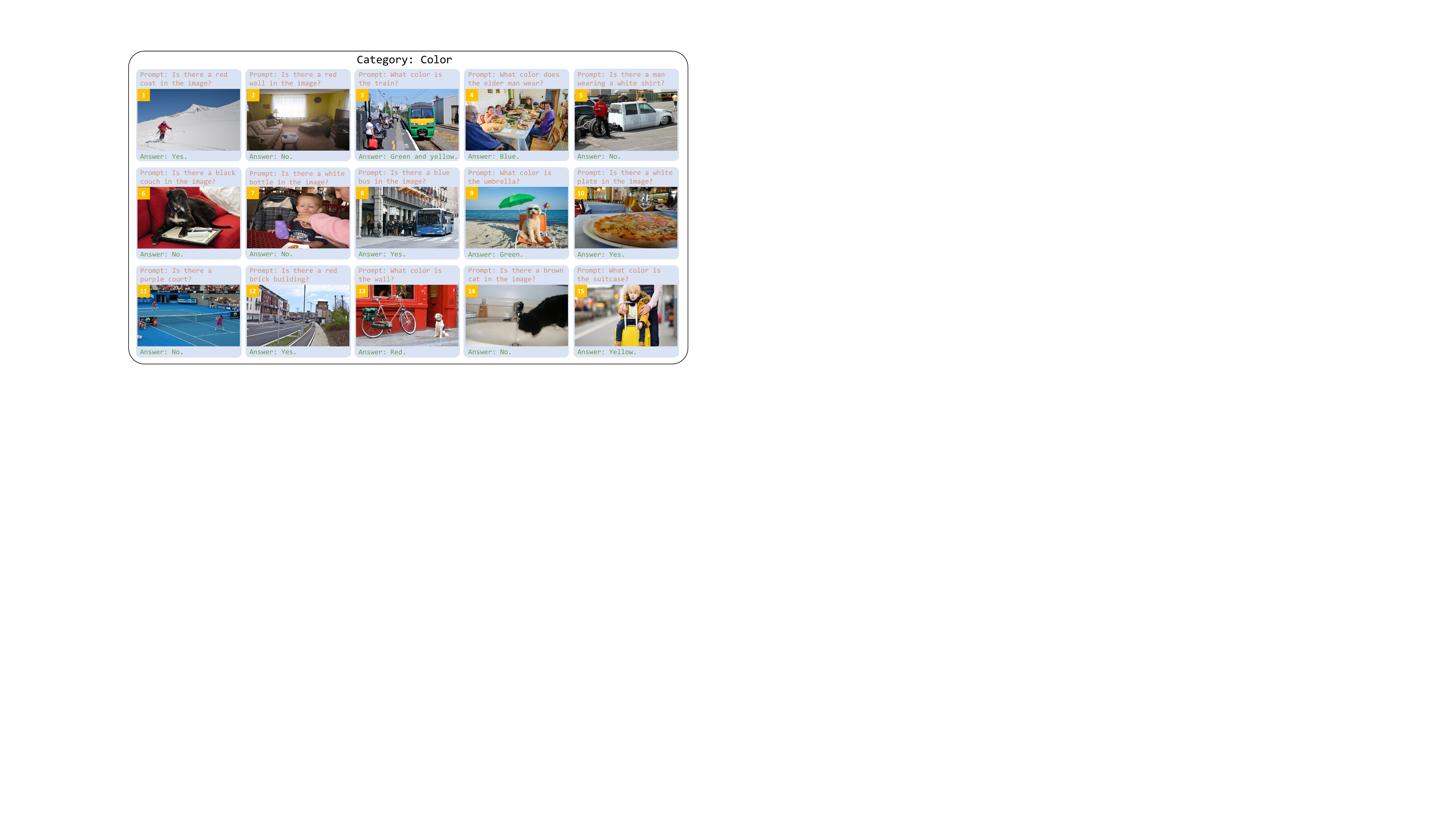}
\caption{Zero-shot image-to-text generation using an MLLM based on the prompt-aware adapter, where it shows favorable visual perception ability on the color recognition task.}
\label{fig:app-color}
\end{figure}

\begin{figure}[h]
\centering
\includegraphics[width=\textwidth]{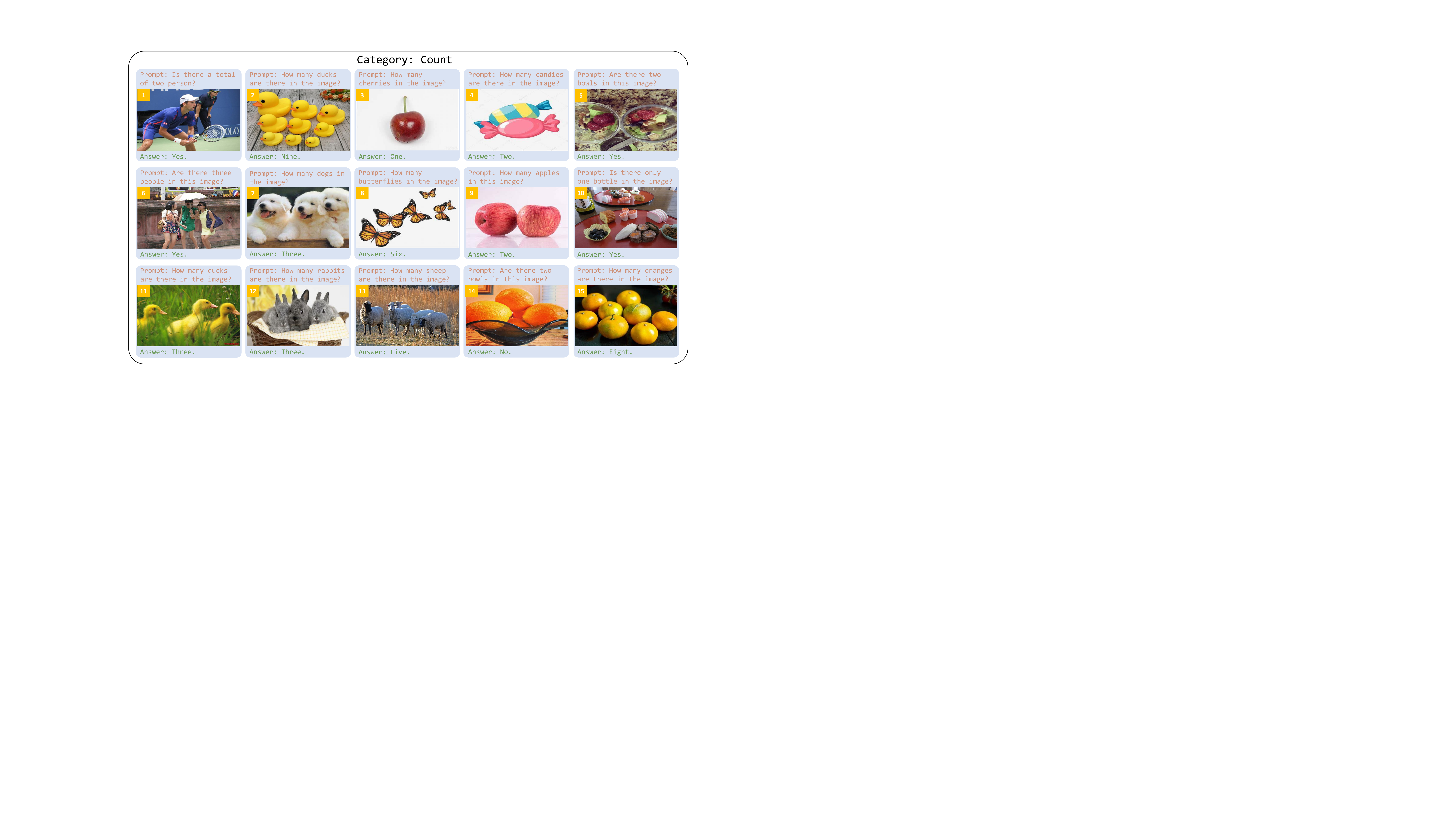}
\caption{Zero-shot image-to-text generation using an MLLM based on the prompt-aware adapter, where it shows favorable visual perception ability on the counting task.}
\label{fig:app-count}
\end{figure}

\begin{figure}[h]
\centering
\includegraphics[width=\textwidth]{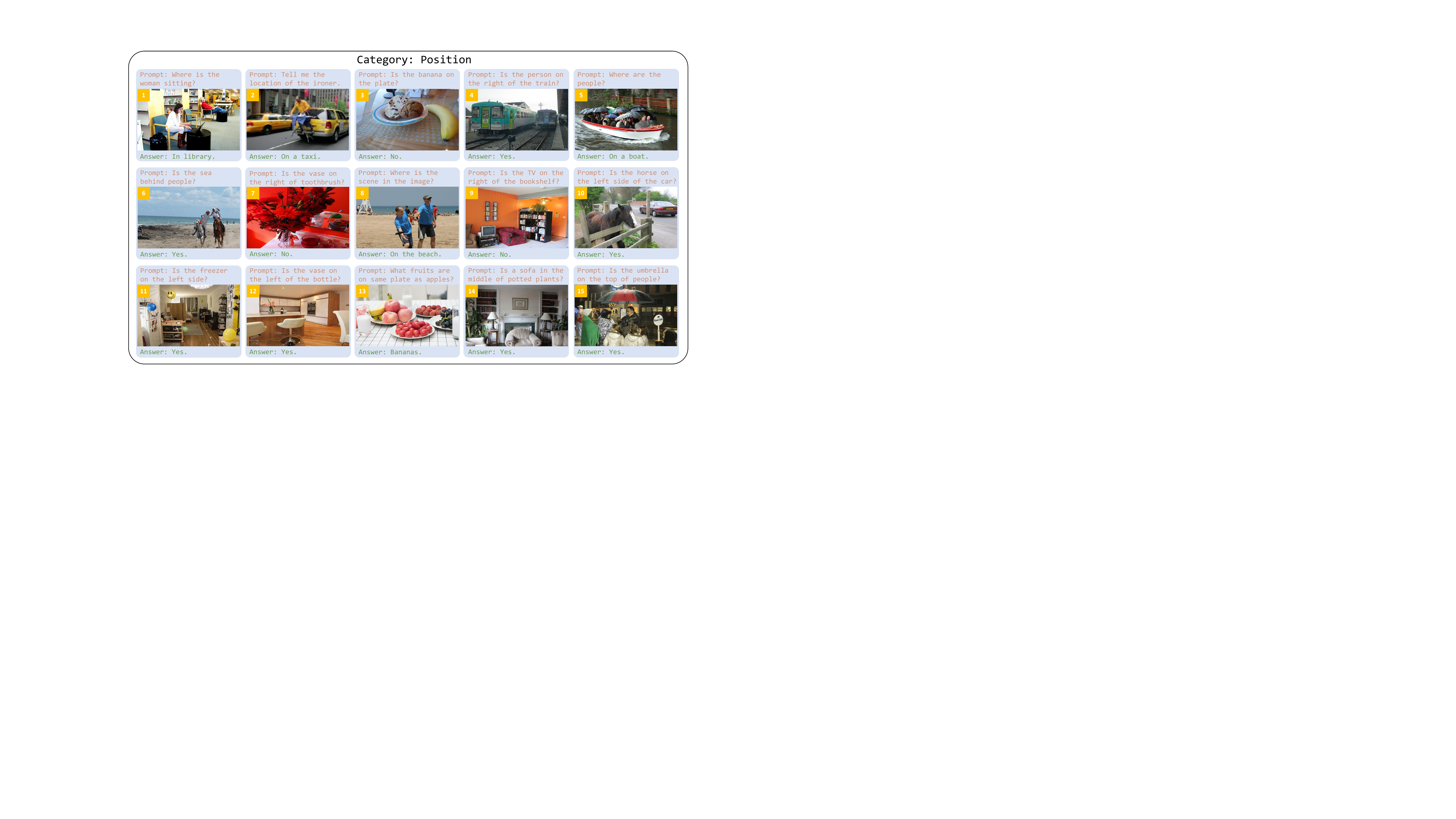}
\caption{Zero-shot image-to-text generation using an MLLM based on the prompt-aware adapter, where it shows favorable visual perception ability on the position reasoning task.}
\label{fig:app-position}
\end{figure}

\begin{figure}[h]
\centering
\includegraphics[width=\textwidth]{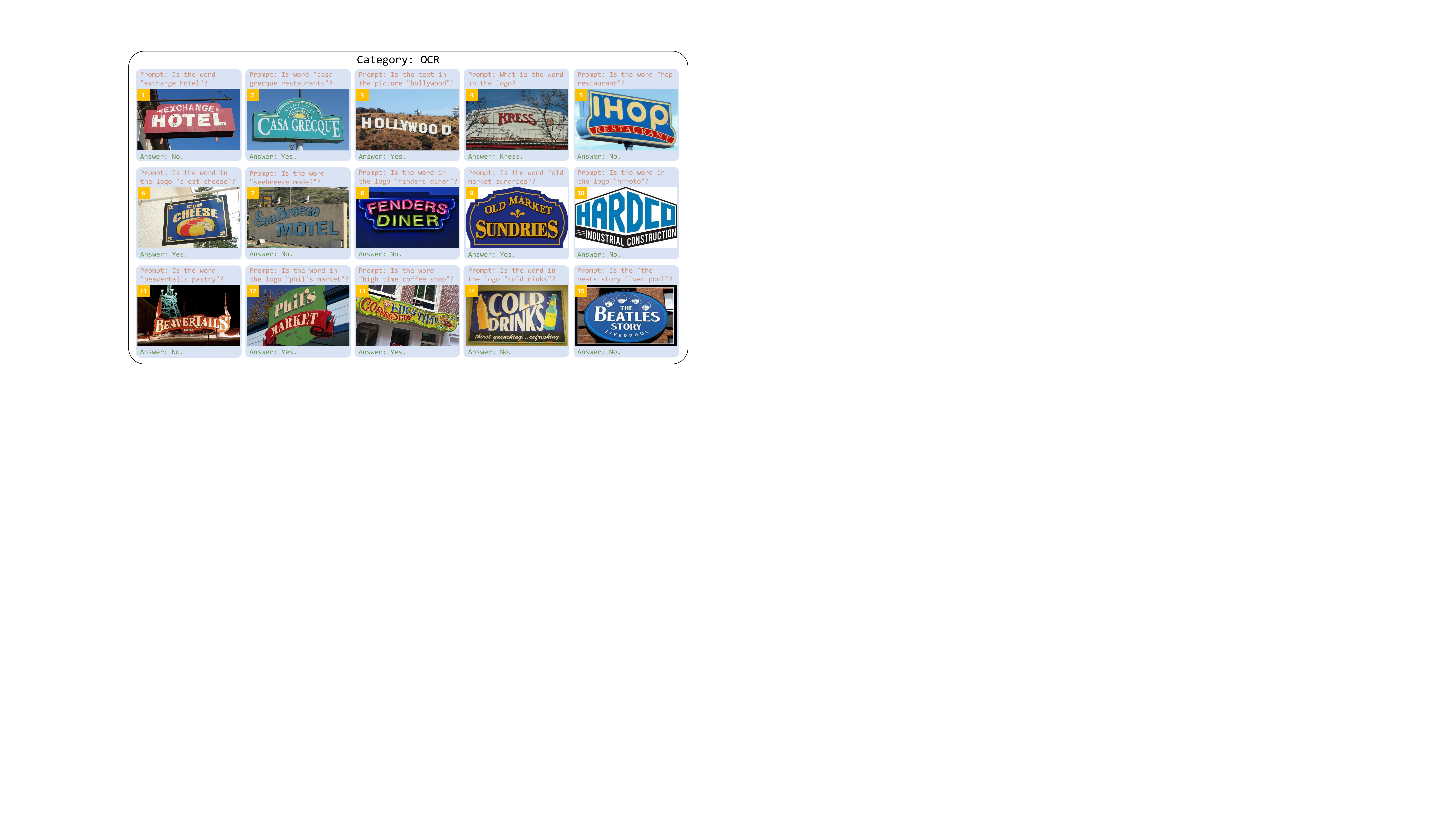}
\caption{Zero-shot image-to-text generation using an MLLM based on the prompt-aware adapter, where it shows favorable visual perception ability on the OCR task.}
\label{fig:app-OCR}
\end{figure}

\begin{figure}[h]
\centering
\includegraphics[width=\textwidth]{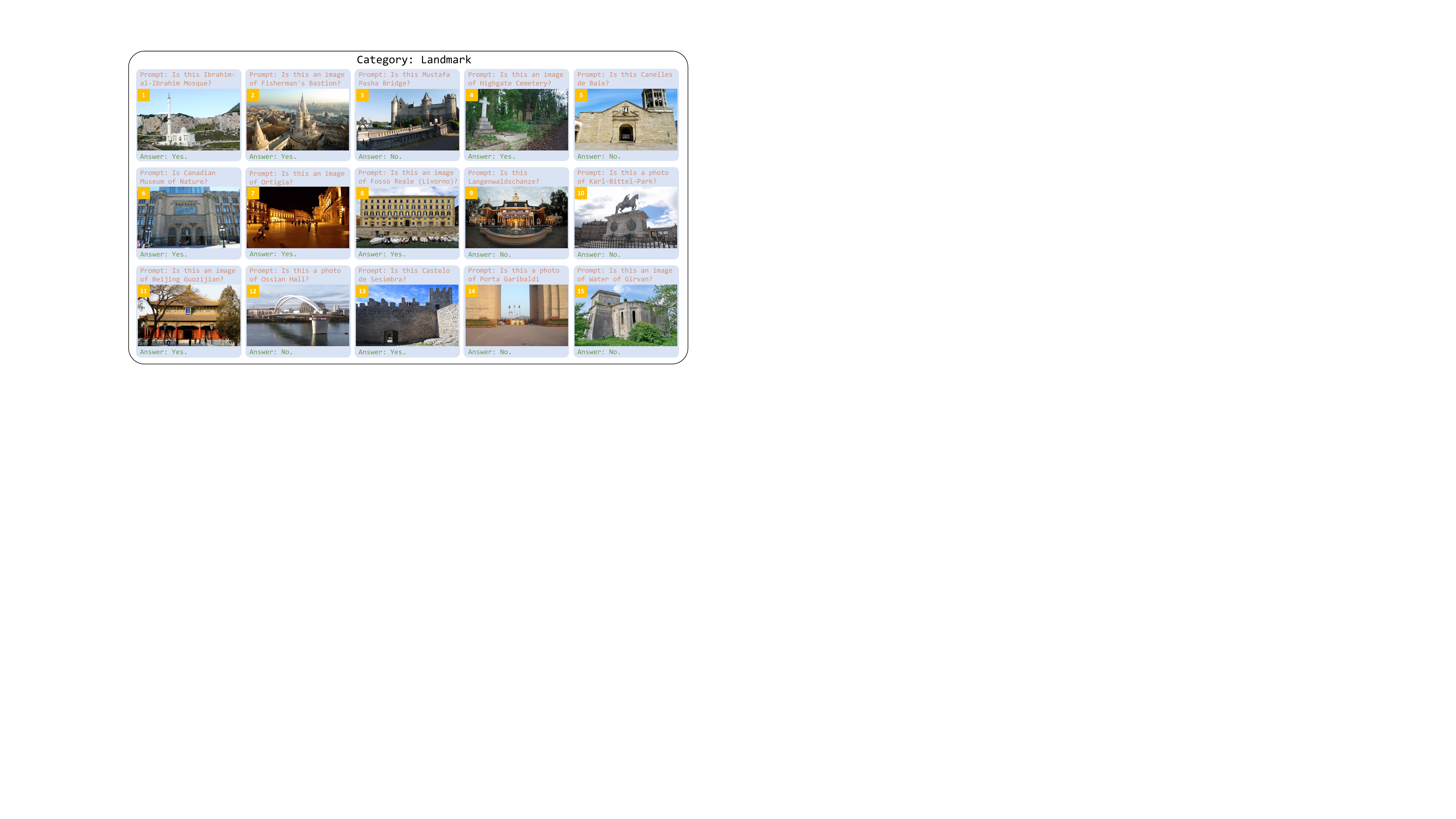}
\caption{Zero-shot image-to-text generation using an MLLM based on the prompt-aware adapter, where it shows favorable visual perception ability on the landmark recognition task.}
\label{fig:app-landmark}
\end{figure}

\begin{figure}[h]
\centering
\includegraphics[width=\textwidth]{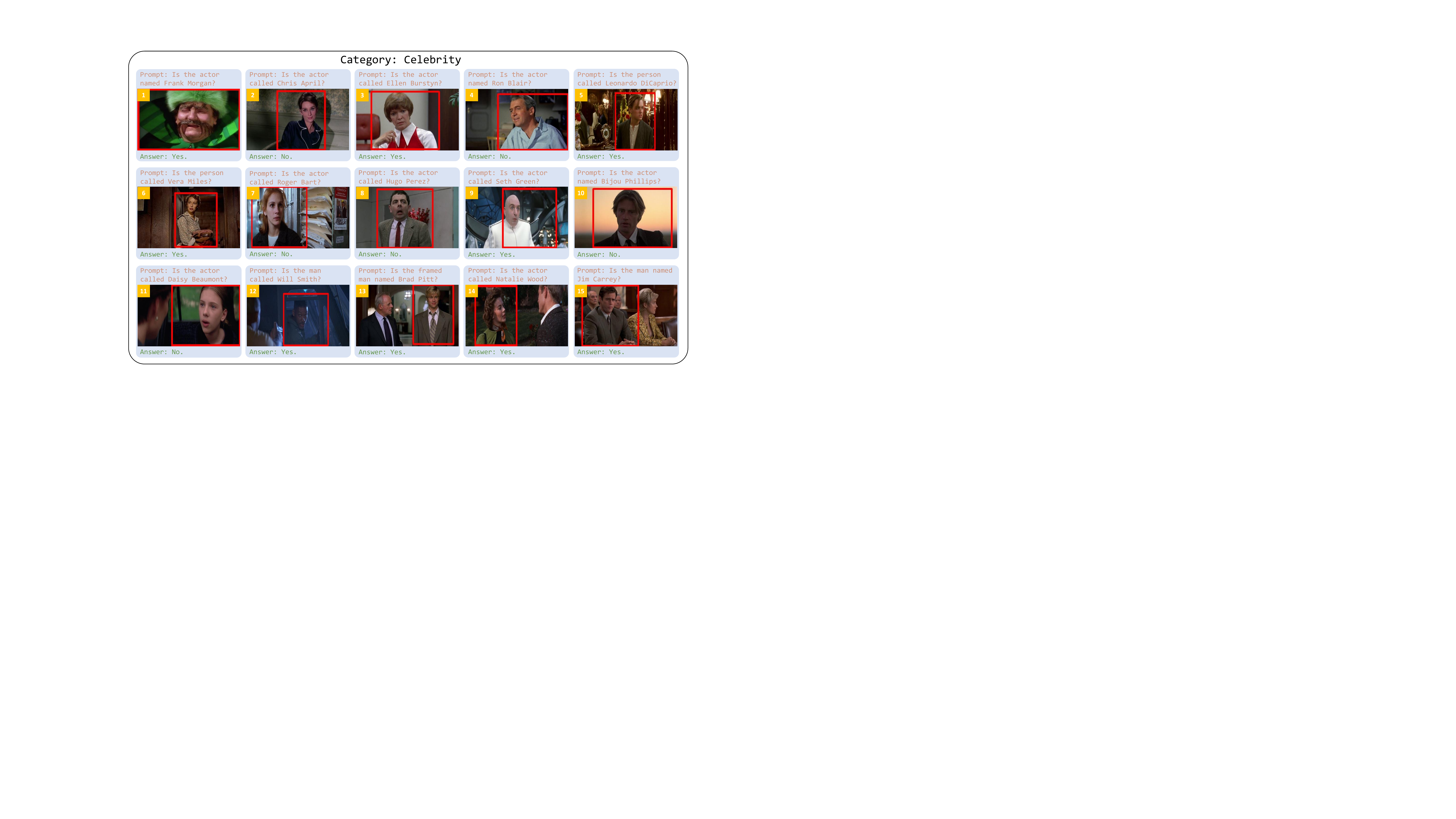}
\caption{Zero-shot image-to-text generation using an MLLM based on the prompt-aware adapter, where it shows favorable visual perception ability on the celebrity recognition task.}
\label{fig:app-celebrity}
\end{figure}

\begin{figure}[h]
\centering
\includegraphics[width=\textwidth]{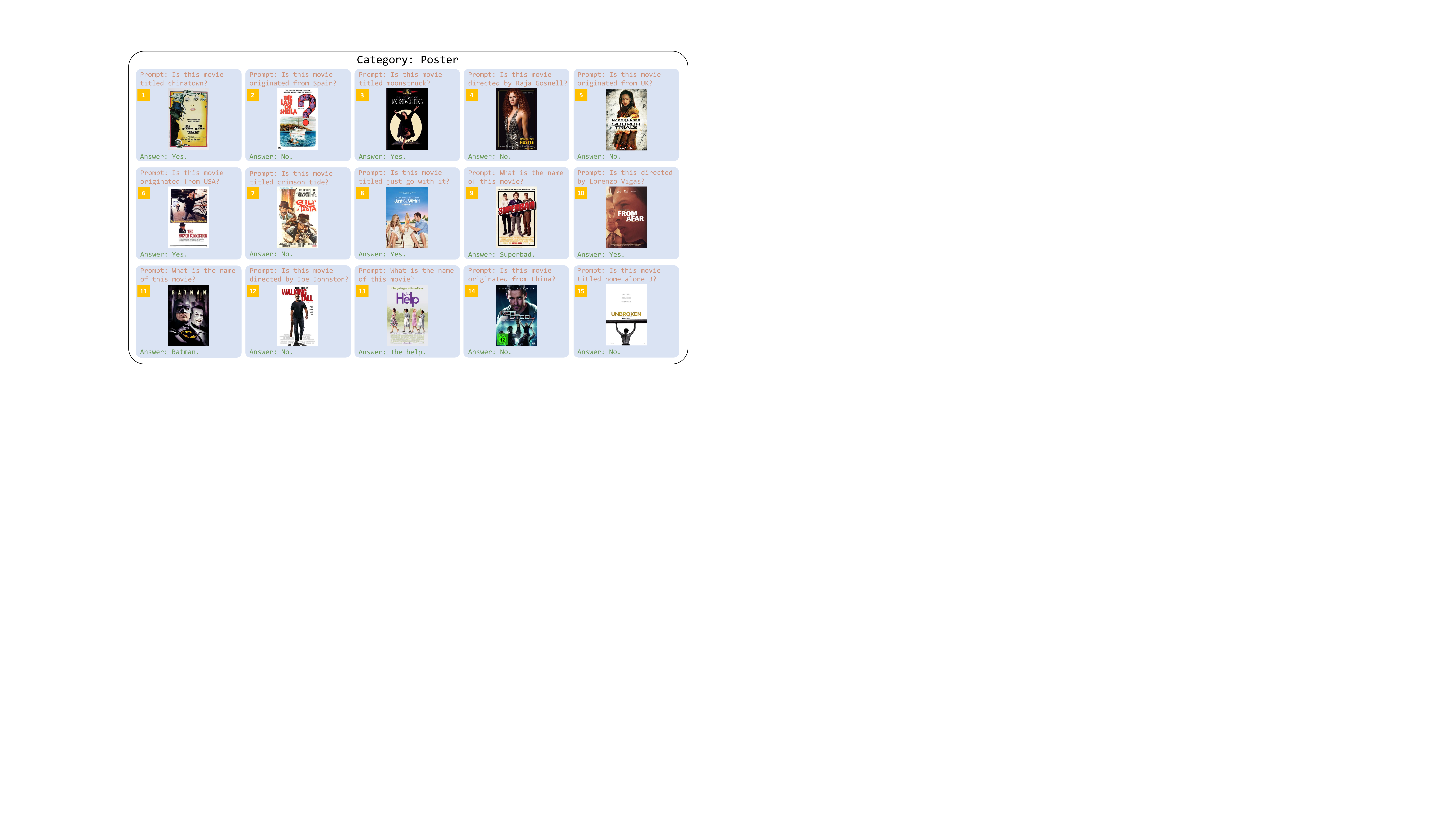}
\caption{Zero-shot image-to-text generation using an MLLM based on the poster recognition task.}
\label{fig:app-poster}
\end{figure}

\begin{figure}[h]
\centering
\includegraphics[width=\textwidth]{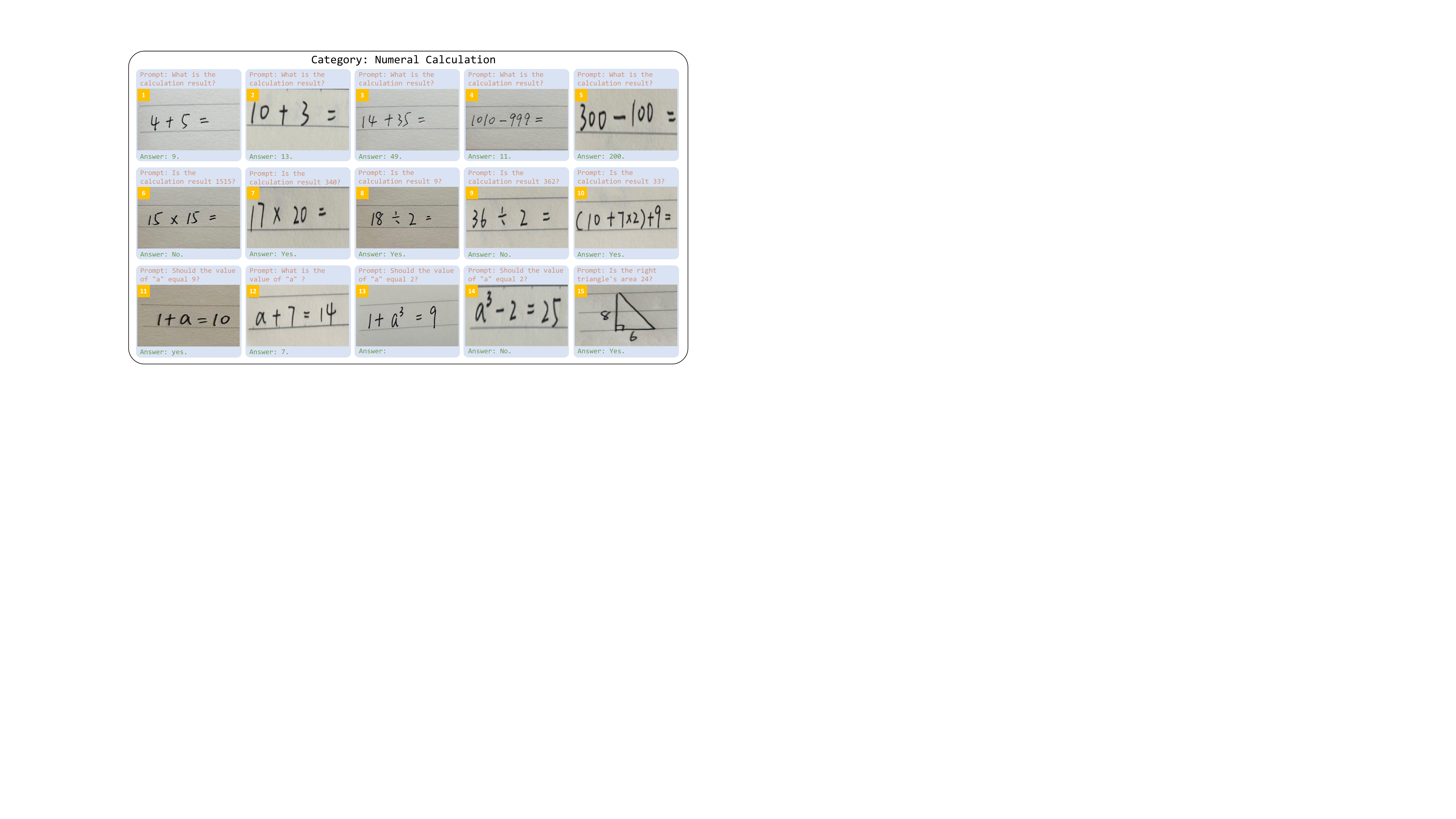}
\caption{Zero-shot image-to-text generation using an MLLM based on the prompt-aware adapter, where it shows favorable visual cognition ability on the calculation task.}
\label{fig:app-calculation}
\end{figure}

\begin{figure}[h]
\centering
\includegraphics[width=\textwidth]{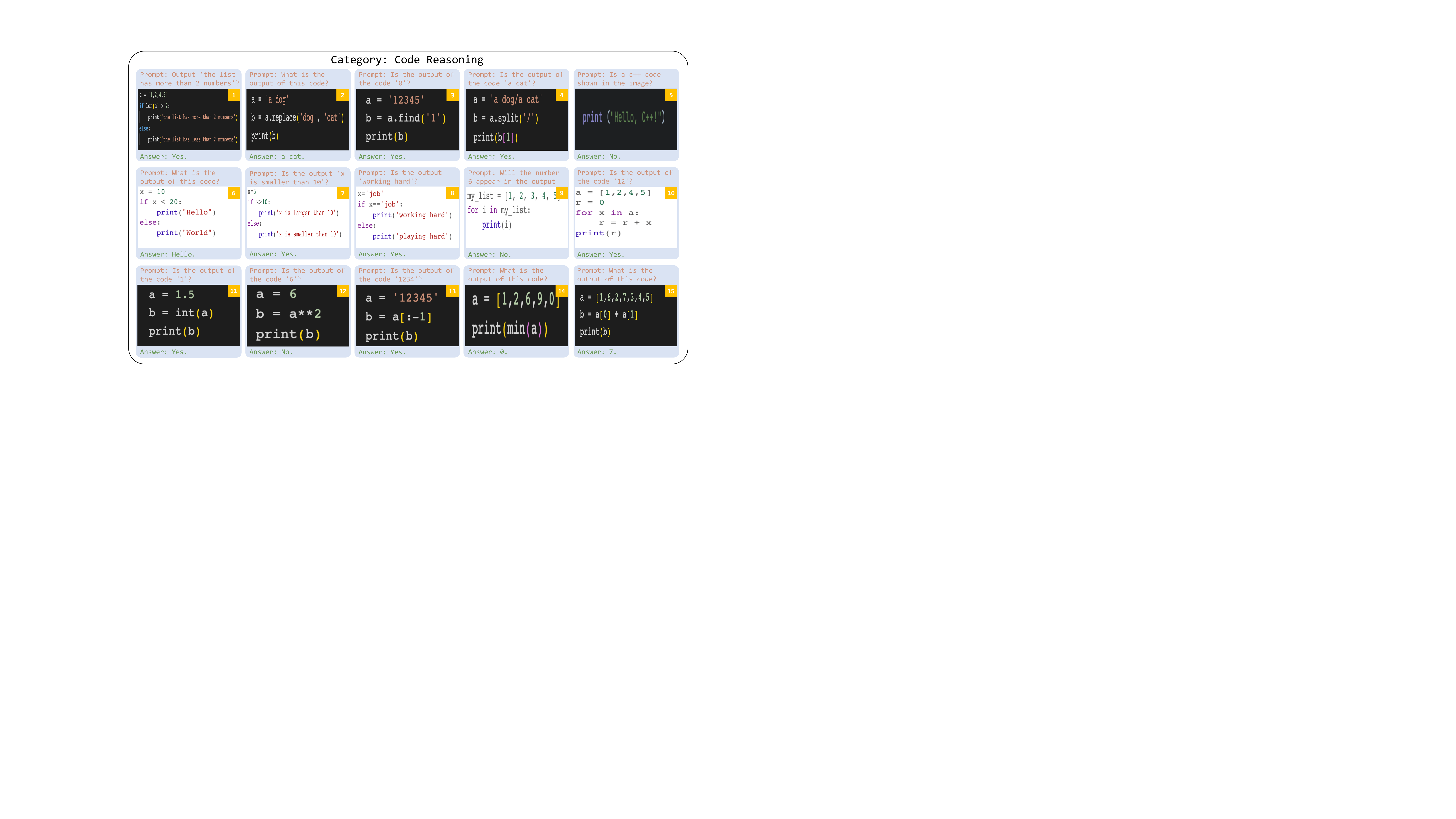}
\caption{Zero-shot image-to-text generation using an MLLM based on the prompt-aware adapter, where it shows favorable visual cognition ability on the code reasoning task.}
\label{fig:app-code}
\end{figure}

\begin{figure}[h]
\centering
\includegraphics[width=\textwidth]{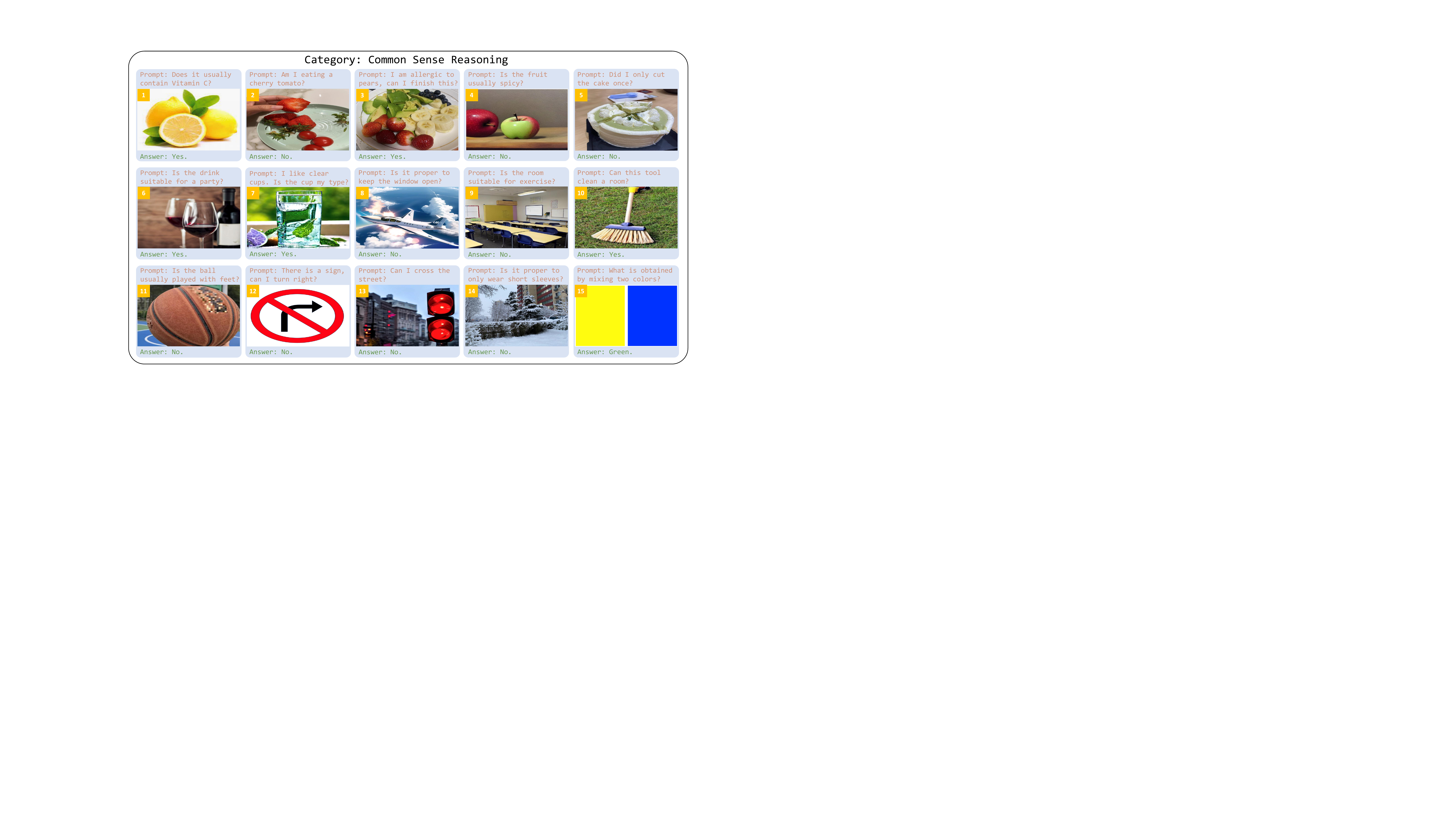}
\caption{Zero-shot image-to-text generation using an MLLM based on the prompt-aware adapter, where it shows favorable visual cognition ability on the commonsense reasoning task.}
\label{fig:app-commensence}
\end{figure}

\begin{figure}[h]
\centering
\includegraphics[width=\textwidth]{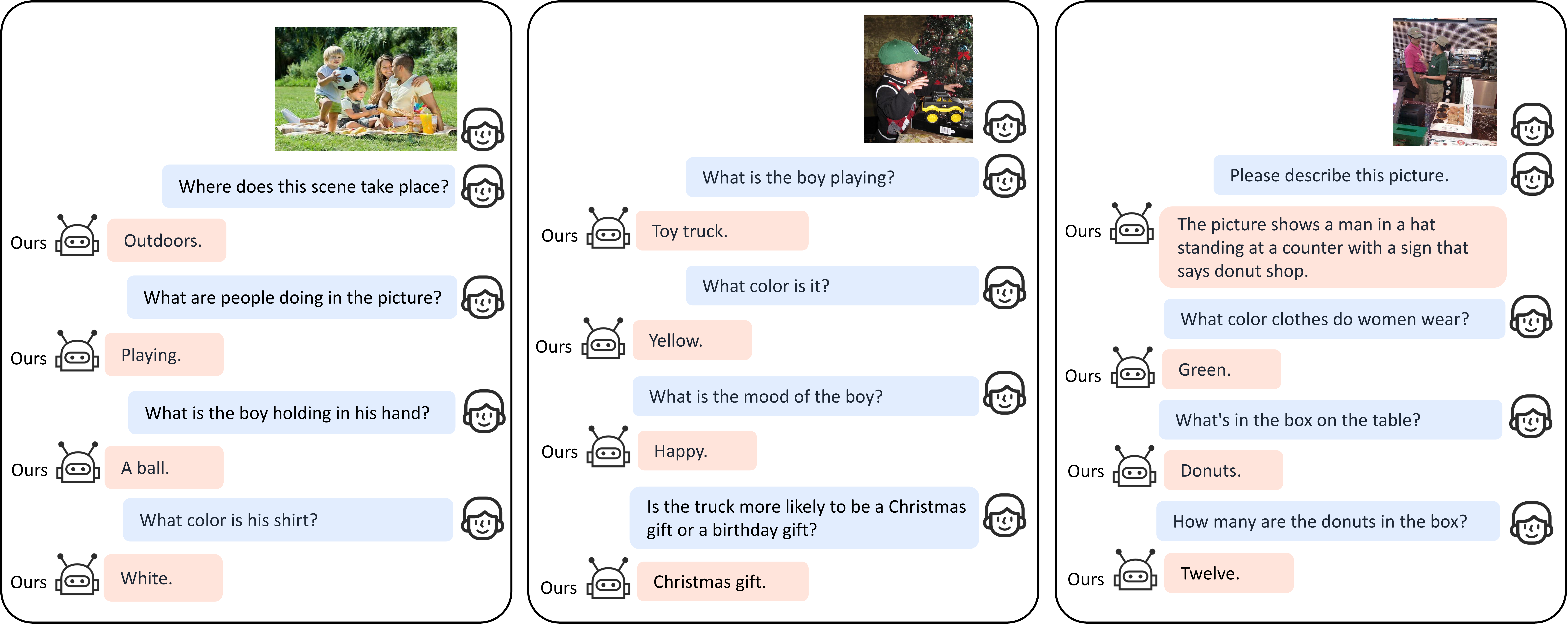}
\caption{Zero-shot image-to-text generation using an MLLM based on the prompt-aware adapter, where it shows favorable visual perception ability.}
\label{fig:app-perception-mix-1}
\end{figure}

\begin{figure}[h]
\centering
\includegraphics[width=\textwidth]{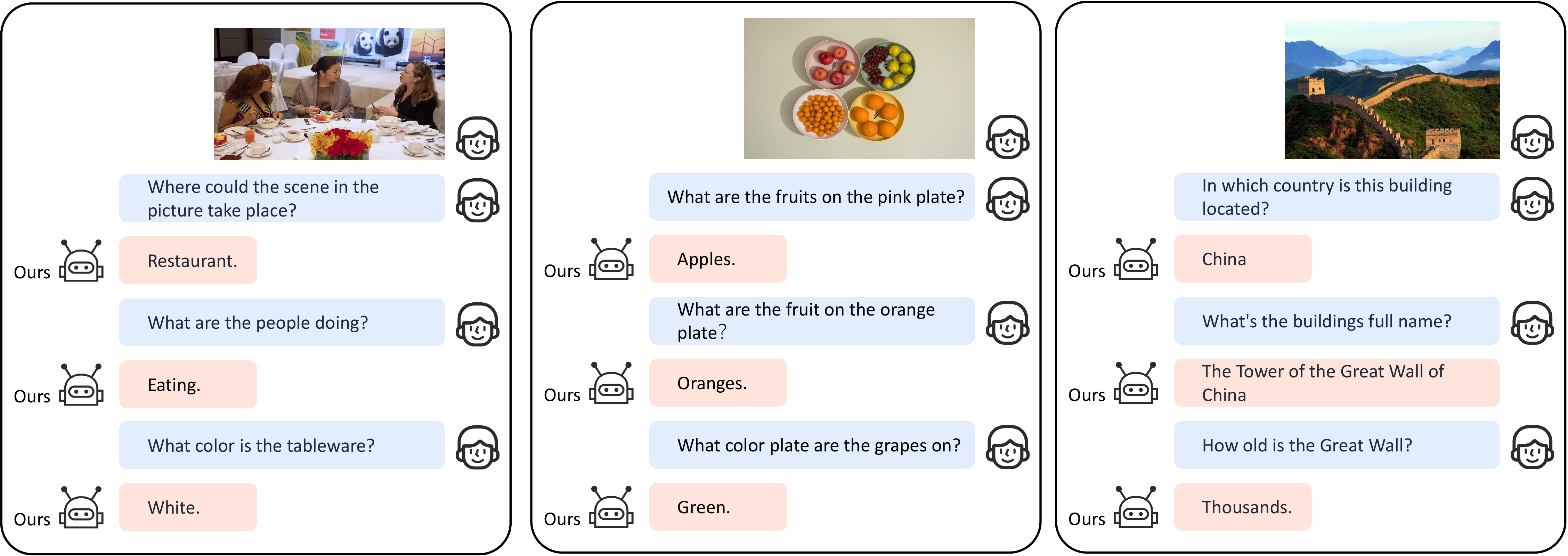}
\caption{Zero-shot image-to-text generation using an MLLM based on the prompt-aware adapter, where it shows favorable visual perception ability.}
\label{fig:app-perception-mix-3}
\end{figure}

We also provide more qualitative results with a wider range of image and prompt inputs (as shown in Fig.s\ref{fig:app-color}$\sim$\ref{fig:app-perception-mix-3}). Specifically, Fig.~\ref{fig:app-color} focuses on the color recognition task, Fig.~\ref{fig:app-count} on the counting task, Fig.~\ref{fig:app-position} on the position reasoning task, Fig.~
\ref{fig:app-OCR} on the OCR task, Fig.~\ref{fig:app-landmark} on the landmark recognition task, Fig.~\ref{fig:app-celebrity} on the celebrity recognition task, and Fig.~\ref{fig:app-poster} on the poster recognition task. Additionally, Fig.~\ref{fig:app-calculation}, Fig.~\ref{fig:app-code}and Fig.~\ref{fig:app-commensence} respectively showcase the model's capabilities in computation, coding, and common sense reasoning. Finally, Fig.s~\ref{fig:app-perception-mix-1}$\sim$\ref{fig:app-perception-mix-3} provide visualization results for mixed perception tasks.

\begin{figure}[h]
\centering
\includegraphics[width=\textwidth]{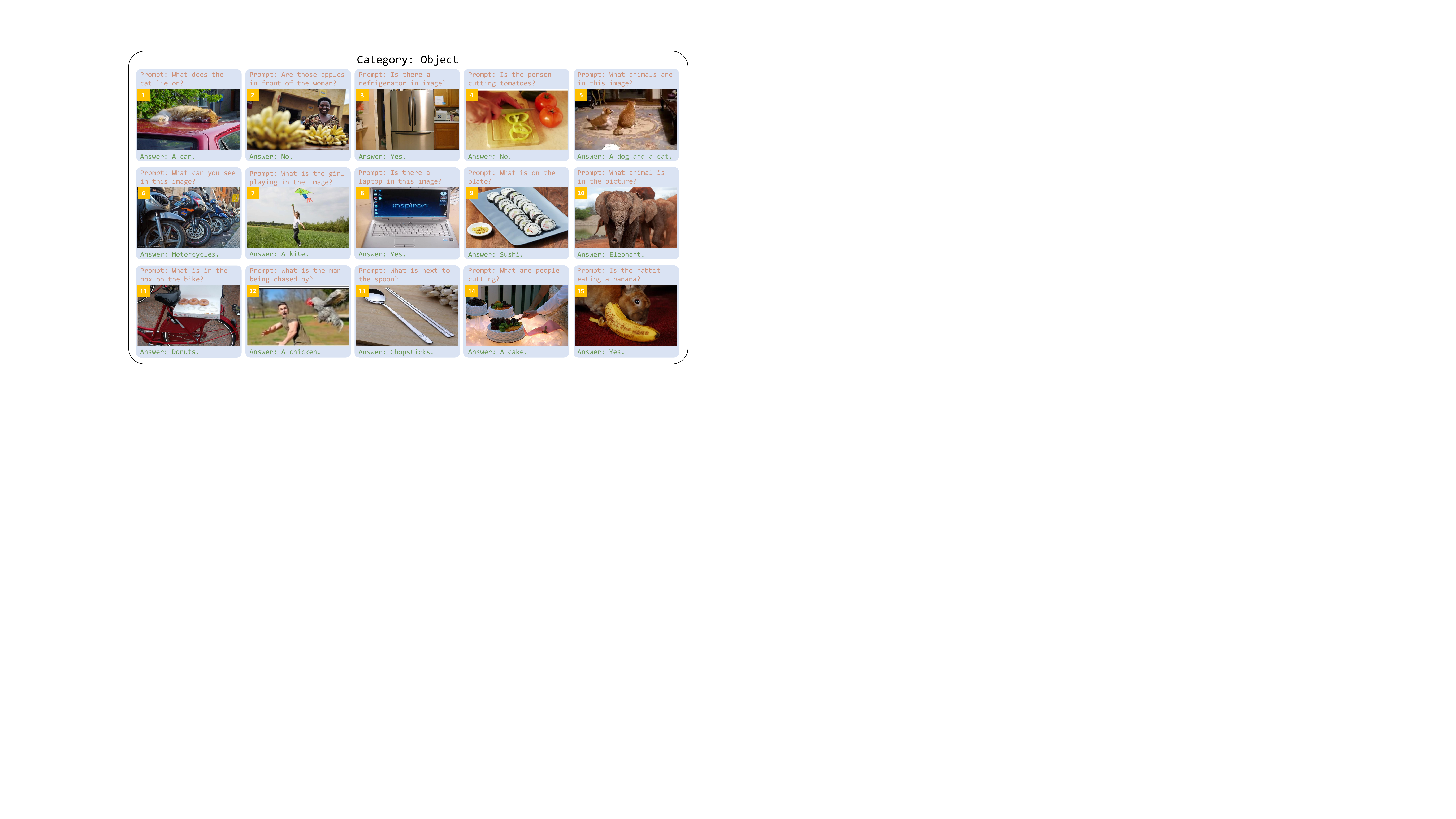}
\caption{Zero-shot image-to-text generation using an MLLM based on the prompt-aware adapter, where it shows favorable visual perception ability on the object detection task.}
\label{fig:app-object}
\end{figure}

\begin{figure}[h]
\centering
\includegraphics[width=\textwidth]{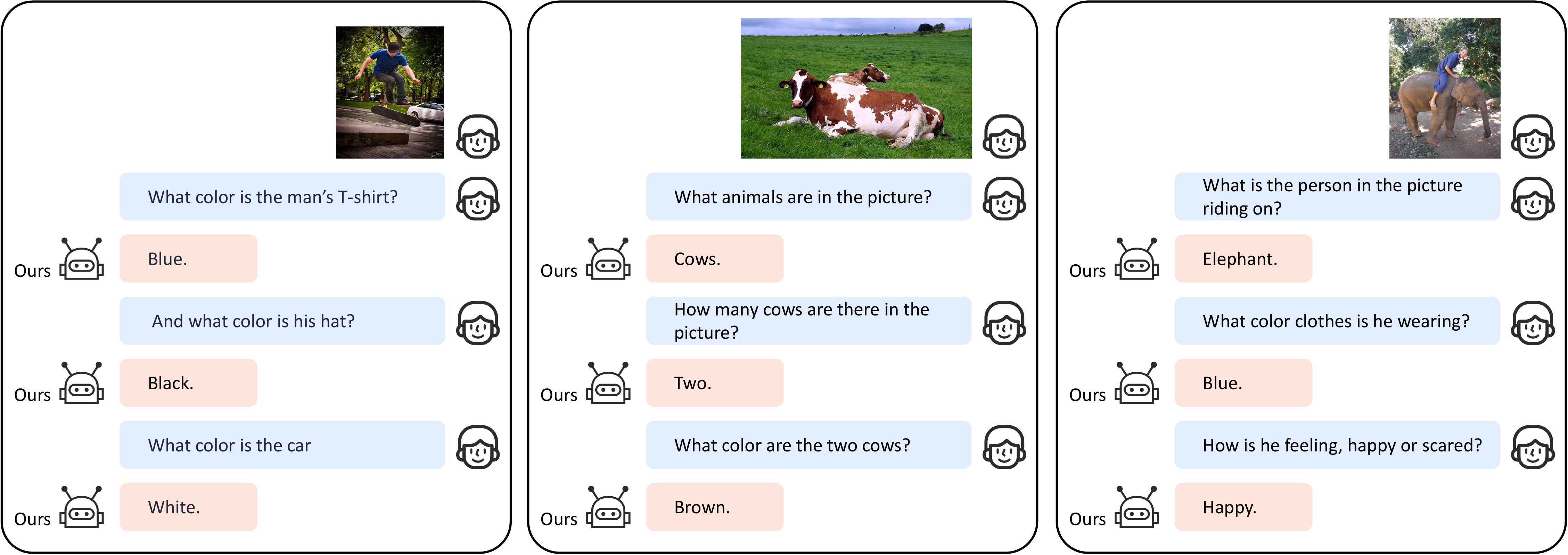}
\caption{Zero-shot image-to-text generation using an MLLM based on the prompt-aware adapter, where it shows favorable visual perception ability.}
\label{fig:app-perception-mix-2}
\end{figure}

\section{More Details of COCO-QA Dataset and MME Benchmark}\label{sec:app-data}
\subsection{COCO-QA Dataset}
    \textbf{Object Classification.}
    The primary objective of the object classification task is to accurately recognize objects in the given visual input and determine the category to which they belong.
    The metric used to evaluate this task is accuracy.
    For this task, we utilize $82,198$ question-answer pairs related to objects from the COCO-QA dataset~\cite{ren2015exploring}, with $78,666$ used for fine-tuning and $3,532$ for testing.  
    The questions typically involve the use of ``what" to inquire about the object type. In addition, a succinct single-word answer is employed to precisely indicate the response.
   
    \textbf{Counting.}
    The main goal of counting is to recognize and count the number of questioned objects from the given visual input.
    The metric employed to assess this task is accuracy.
    In the counting task, we use $19,568$ question-answer pairs related to the number of objects from the COCO-QA dataset~\cite{ren2015exploring}, with $1,8761$ used for fine-tuning and $807$ for testing.  
    When posing questions, the inquiry typically involves the use of ``how many" to inquire about the object amount. Likewise, we only use a single word describing the quantity in the response.
    
    \textbf{Color Recognition.}
    The color recognition task is designed to detect objects in questions and perceive color information from the input visual signal.
    The metric employed to assess this task is also accuracy.
    In this task, we utilize $8,640$ question-answer pairs related to objects from the COCO-QA dataset~\cite{ren2015exploring}, with $8,304$ used for fine-tuning and $336$ for testing. 
    Questions related to color are relatively straightforward, usually beginning with ``What is the color of".
     To facilitate quantitative evaluation, responses responses are kept as brief as possible for color-related words.

    \textbf{Position Reasoning.}
    In the position reasoning task, our primary goal is to infer the location information of the queried object based on the input visual content.
    Similarly, we adopt the accuracy as the evaluation metric.
    We adopt $7,278$ question-answer pairs related to objects from the COCO-QA dataset~\cite{ren2015exploring}, with $6,953$ used for fine-tuning and $325$ for testing. 
    In this task, questions are formatted to start with ``where" to ask for the position of an object.
    Answers mostly be places, scenes, or large objects that contain smaller objects.

Examples of questions and answers from four perceptual tasks are shown in Table~\ref{tab:app-question}.
\begin{table}[h]
\centering
\caption{Examples of questions and answers from four visual perception tasks.}
\resizebox{\textwidth}{!}{
    \begin{tabular}{l|l|l} 
    \toprule
      Visual Perception Task & Question & Answer\\
    \midrule
    \multirow{5}*{Object Classification} & \textit{what are sitting down on the ground}& \textit{bears} \\
       & \textit{what is parked on the side of the grass} &\textit{motorcycle}\\
       & \textit{what are two mhttps://www.overleaf.com/project/64eb0aa69f2b1a658d35a15ben playing with some elephants} &\textit{ball}\\
       & \textit{what is the color of the shirt} &\textit{blue}\\
       & \textit{what is laying on the bed next to some pillows} &\textit{cat}\\
    \midrule
    \multirow{5}*{Counting} & \textit{how many men is sitting on the street in front of a building} & \textit{two}\\
       & \textit{how many red velvet cup cakes with no frosting on a flowered plate} &\textit{three}\\
       & \textit{how many pairs of shoes on a mat with a cat is sitting in the middle} &\textit{eight}\\
       & \textit{how many dessert treats in the white cardboard box} &\textit{six}\\
       & \textit{how many trays of itallian food are in large pans} &\textit{four}\\
    \midrule
    \multirow{5}*{Color Recognition} & \textit{what is the color of the airplane}& \textit{black}\\
       & \textit{what is the color of the motorcycle} &\textit{orange}\\
       & \textit{what is the color of the brush} &\textit{green}\\
       & \textit{what is the color of the bird} &\textit{white}\\
       & \textit{what is the color of the flowers} &\textit{red}\\
    \midrule
    \multirow{5}*{Position Reasoning} & \textit{where is the cat lounging} &\textit{chair}\\
       & \textit{where do the mother and son make sundaes} &\textit{kitchen}\\
       & \textit{where is the cheese pizza} &\textit{box}\\
       & \textit{where is the person sitting} &\textit{bed}\\
       & \textit{where do the large and over-sized stuffed teddy bear sitting} &\textit{chair}\\    
        \bottomrule        
    \end{tabular}
    }   
    \label{tab:app-question}
\end{table}

\subsection{MME Benckmark}
MME~\cite{fu2023mme} benchmark evaluates both the perception and cognition abilities of MLLMs. In addition to OCR, perception encompasses the recognition of both coarse-grained and fine-grained objects. Coarse-grained recognition includes the identification of objects' existence, count, position, and color. Fine-grained recognition involves identifying movie posters, celebrities, scenes, landmarks, and artworks. Cognition tasks include commonsense reasoning, numerical calculation, text translation, and code reasoning. There are a total of 14 sub-tasks, detailed as follows.

\textbf{Coarse-Grained Recognition.}
The contents of coarse-grained recognition include identifying the existence of common objects and determining their count, color, and position. Images are sourced from the COCO dataset~\cite{lin2014microsoft}, but all instruction-answer pairs are manually constructed rather than directly using available annotations. Even if MLLMs have previously encountered these COCO images, the manually prepared pairs are not included in their training sets. This setup requires MLLMs to comprehend the instructions and infer the correct answers. For each perception subtask of existence, count, color, and position, 30 images are used, with a total of 60 instruction-answer pairs prepared.

\textbf{Fine-Grained Recognition.}
Fine-grained recognition evalutates the knowledge resources of MLLMs, encompassing subtasks such as recognizing movie posters, celebrities, scenes, landmarks, and artworks. These subtasks include 147, 170, 200, 200, and 200 images respectively. For the celebrity subtask, a red box is placed around a person with a clearly visible face in the image, with the instruction being, “Is the actor inside the red box named [celebrity name]? Please answer yes or no.” Similar to the coarse-grained recognition, images for these subtasks are sourced from publicly available datasets~\cite{DBLP:journals/corr/abs-2007-10937, DBLP:conf/mm/MaoCS17, DBLP:journals/tomccap/MaoSC19, DBLP:conf/cvpr/WeyandACS20, DBLP:conf/nips/ZhouLXTO14}, and all instructions are manually crafted.

\textbf{Optical Character Recognition.}
Optical Character Recognition (OCR) is a foundational capability of MLLMs, supporting subsequent text-based tasks such as text translation and text understanding. Images are sampled from~\cite{DBLP:journals/pr/LiuJZLZ19}, with all instruction-answer pairs manually designed. Given that MLLMs are still in their early stages, only relatively simple samples are chosen for this version of MME. This section includes 20 images and 40 instruction-answer pairs.

All image-prompt-answer pairs are manually created. The few public datasets used in this benchmark utilize only the images without relying on their original annotations.
Additionally, images are collected through real photographs and image generation.
The prompts in MME are designed to be concise to minimize the influence of prompt engineering on the model's output.
Due to the instruction design of ``please answer yes or no'' quantitative statistics can be easily performed based on the ``yes'' or ``no'' responses from MLLMs, ensuring accuracy and objectivity.

\section{Prompts used for Fine-tuning}\label{sec:app-prompt}
The prompt templates employed in the end-to-end prompt fine-tuning process across various tasks are illustrated in Table~\ref{tab:app-prompt-template}.
\begin{table}[h]
\centering
\caption{Prompt templates used for converting datasets into prompt-tuning data. In VQA tasks, two formats of prompts are available. Note that the ``\{Question\}" format is mandatory for the prompt-aware adapter, and needs to be replaced with the specific question in Table~\ref{tab:app-question}.}
\resizebox{\textwidth}{!}{
    \begin{tabular}{l|l} 
    \toprule
      Task & Prompt Template\\
    \midrule
    \multirow{5}*{Image Captioning} &         \textit{[INST]~\textless Img\textgreater~\textless Image Feature\textgreater ~ \textless/Img\textgreater~[Caption]~\textless Describe this image in detail.\textgreater} \textit{[/INST]\qquad} \\ 
        & \textit{[INST]~\textless Img\textgreater~\textless Image Feature\textgreater ~ \textless/Img\textgreater~[Caption]~\textless Take a look at this image and describe what you notice.\textgreater} \textit{[/INST]\qquad} \\ 
        & \textit{[INST]~\textless Img\textgreater~\textless Image Feature\textgreater ~ \textless/Img\textgreater~[Caption]~\textless Please provide a detailed description of the picture.\textgreater} \textit{[/INST]\qquad} \\
        & \textit{[INST]~\textless Img\textgreater~\textless Image Feature\textgreater ~ \textless/Img\textgreater~[Caption]~\textless Could you describe the contents of this image for me?\textgreater} \textit{[/INST]\qquad} \\
        & \textit{[INST]~\textless Img\textgreater~\textless Image Feature\textgreater ~ \textless/Img\textgreater~[Caption]~\textless Is this image common in real world?\textgreater} \textit{[/INST]\qquad} \\
    \midrule
    \multirow{6}*{Visual Question Answering} & 
     \textit{[INST]~\textless Img\textgreater~\textless Image Feature\textgreater ~ \textless/Img\textgreater~[vqa]~\textless Describe this image in detail. \textgreater} \textit{[/INST]\qquad} \\
     & \textit{[INST]~\textless Img\textgreater~\textless Image Feature\textgreater ~ \textless/Img\textgreater~[vqa]~\textless Take a look at this image and describe what you notice. \textgreater} \textit{[/INST]\qquad} \\
     & \textit{[INST]~\textless Img\textgreater~\textless Image Feature\textgreater ~ \textless/Img\textgreater~[vqa]~\textless Could you describe the contents of this image for me? \textgreater} \textit{[/INST]\qquad} \\
     & \textit{[INST]~\textless Img\textgreater~\textless Image Feature\textgreater ~ \textless/Img\textgreater~[vqa]~\textless Is this image common in real world? \textgreater} \textit{[/INST]\qquad} \\ 
     & \textit{[INST]~\textless Img\textgreater~\textless Image Feature\textgreater ~ \textless/Img\textgreater~[vqa]~\textless Question \textgreater} \textit{[/INST]\qquad} \\

        \bottomrule        
    \end{tabular}
    }
    \label{tab:app-prompt-template}
\end{table}

\end{document}